%% file: main_arxiv.tex
\documentclass{article}
\usepackage{graphicx} 
\usepackage[preprint]{neurips_2026}
\usepackage[utf8]{inputenc} 
\usepackage[T1]{fontenc}    
\usepackage{hyperref}       
\usepackage{url}            
\usepackage{booktabs}       
\usepackage{amsfonts}       
\usepackage{nicefrac}       
\usepackage{microtype}      
\usepackage{enumitem} 
\usepackage{wrapfig}
\usepackage{xspace}
\usepackage{multirow}
\usepackage{makecell}
\usepackage{threeparttable}
\usepackage{array}
\usepackage[normalem]{ulem}
\usepackage[table]{xcolor}
\usepackage{subcaption}

\newcommand{\score}[2]{#1{\scriptsize$\pm$#2}}

\DeclareRobustCommand{\model}{\textsc{MasFACT}\xspace}

\definecolor{masfactgreen}{HTML}{087F5B}
\definecolor{masfactrow}{HTML}{EEF8F4}

\newcommand{\plugup}[2]{\textbf{#1}{\scriptsize\,\textcolor{masfactgreen}{$\uparrow$#2}}}
\newcommand{\plugdown}[2]{\textbf{#1}{\scriptsize\,\textcolor{masfactgreen}{$\downarrow$#2}}}

\usepackage[authormarkup=none]{changes}
\definechangesauthor[name={wangxf}, color=blue]{wxf}

\newcommand{\nop}[1]{}

\usepackage{amsmath}
\usepackage{amssymb}

\usepackage{amsthm}
\newtheorem{theorem}{Theorem}

\usepackage{algorithm}
\usepackage{algorithmic}
\usepackage{newtxtext}

\input{math_commands}

\newcommand{\authorentry}[3]{%
\begin{tabular}[t]{@{}c@{}}
\textbf{#1}\\
#2\\
\texttt{#3}
\end{tabular}%
}

\newcommand{\correspondingauthornote}{%
\begingroup
\renewcommand{\thefootnote}{}%
\footnotetext{\textsuperscript{*}Corresponding author.}%
\addtocounter{footnote}{-1}%
\endgroup
}

\title{\textsc{MasFACT}: Continual Multi-Agent Topology Learning via Geometry-Aware Posterior Transfer}

\author{%
\normalfont\small
\begin{tabular}{@{}c@{\hspace{1.4em}}c@{\hspace{1.4em}}c@{}}
\authorentry{Xuefei Wang}{Beihang University}{xuefeiw@buaa.edu.cn}
&
\authorentry{Jialu Wang}{Independent Researcher}{faldict@ucsc.edu}
&
\authorentry{Fengbo Zhang}{Beijing University of Posts and\\Telecommunications}{fengboz@bupt.edu.cn}
\\[2.2ex]
\authorentry{Yihan Hu}{Beihang University}{yihan\_hu@buaa.edu.cn}
&
\authorentry{Di Zhang}{Beihang University}{imzhangdi@buaa.edu.cn}
&
\authorentry{Yutong Ye}{Beihang University}{yutongye@buaa.edu.cn}
\\[2.2ex]
\authorentry{Yikun Ban}{Beihang University}{yikunb@buaa.edu.cn}
&
\authorentry{Jun Han}{Beihang University}{jun\_han@buaa.edu.cn}
&
\authorentry{Ruijie Wang\textsuperscript{*}}{Beihang University}{ruijiew@buaa.edu.cn}
\end{tabular}%
}

\begin{document}

\maketitle
\correspondingauthornote
\vspace{-1.5em}

\begin{abstract}

Multi-agent systems (MAS) powered by large language models (LLMs) have emerged as a powerful paradigm for complex problem solving, where performance critically depends on the underlying inter-agent communication topology. However, existing topology generation methods mainly optimize for isolated tasks, while real-world deployments involve streams of evolving tasks, requiring previously effective collaboration patterns to be retained and reused rather than rediscovered or overwritten. We identify a previously underexplored failure mode, \emph{topology forgetting}, in which adapting to new tasks shifts the topology generator away from communication structures required by earlier tasks. This issue stems from cross-task misalignment in both agent-level functional semantics and relational communication structures. To address this challenge, we propose \textbf{\model}, a geometry-aware posterior transfer framework that preserves and reuses historical collaboration knowledge as transferable topology priors. We transfer these priors across task-specific agent spaces through Fused Gromov-Wasserstein optimal transport and perform PAC-Bayes-guided conservative posterior adaptation to balance task-specific plasticity with structural stability. Experiments across class-, domain-, and task-level continual settings demonstrate that \model consistently improves average accuracy while reducing topology forgetting compared to strong topology generation and replay-based baselines, and can be seamlessly integrated with different MAS topology generators.

\end{abstract}

\input{text/01-introduction}
\input{text/03-definition}
\input{text/04_method_v2}
\input{text/05-experiments}

\input{text/02-related}
\input{text/06-conclusion}

\bibliographystyle{plain}
\bibliography{ref}
\newpage
\appendix

\input{text/09-appendix}

\end{document}

%% file: math_commands.tex
\usepackage{amsmath,amsfonts,bm}

\def\1{\bm{1}}

\def\vone{{\bm{1}}}
\def\vmu{{\bm{\mu}}}

\def\vnu{{\bm{\nu}}}

\def\mB{{\bm{B}}}
\def\mC{{\bm{C}}}

\def\mQ{{\bm{Q}}}
\def\mR{{\bm{R}}}
\def\mS{{\bm{S}}}
\def\mT{{\bm{T}}}

\def\mX{{\bm{X}}}

\def\mZ{{\bm{Z}}}

\DeclareMathAlphabet{\mathsfit}{\encodingdefault}{\sfdefault}{m}{sl}
\SetMathAlphabet{\mathsfit}{bold}{\encodingdefault}{\sfdefault}{bx}{n}
\newcommand{\tens}[1]{\bm{\mathsfit{#1}}}

\def\tC{{\tens{C}}}

\def\gA{{\mathcal{A}}}
\def\gB{{\mathcal{B}}}

\def\gD{{\mathcal{D}}}

\def\gG{{\mathcal{G}}}
\def\gH{{\mathcal{H}}}

\def\gM{{\mathcal{M}}}

\def\gR{{\mathcal{R}}}
\def\gS{{\mathcal{S}}}
\def\gT{{\mathcal{T}}}
\def\gU{{\mathcal{U}}}

\newcommand{\E}{\mathbb{E}}
\newcommand{\Ls}{\mathcal{L}}
\newcommand{\R}{\mathbb{R}}

\newcommand{\KL}{D_{\mathrm{KL}}}

\DeclareMathOperator*{\argmax}{arg\,max}
\DeclareMathOperator*{\argmin}{arg\,min}

\newcommand{\eg}{e.g.\ }

%% file: text/01-introduction.tex
\section{Introduction}
\label{introduction}

Large language model (LLM)-based multi-agent systems (MAS) have emerged as a powerful paradigm for complex reasoning, planning, and decision-making tasks~\cite{guo2024large,wu2024autogen}. 
A central factor governing their effectiveness is the underlying communication topology, which determines how agents coordinate, exchange information, and aggregate intermediate reasoning~\cite{du2024improving,shen2025understanding,hong2023metagpt}. 
An effective topology can substantially improve reasoning quality and coordination efficiency while mitigating structural inefficiencies such as redundant communication and context dilution~\cite{fan2026todycomm}. 
Consequently, topology design has become a fundamental bottleneck in scaling reliable and adaptive LLM-based MAS.

\begin{figure}[t]
    \centering
    \begin{subfigure}[t]{0.24\linewidth}
        \centering
        \includegraphics[width=\linewidth]{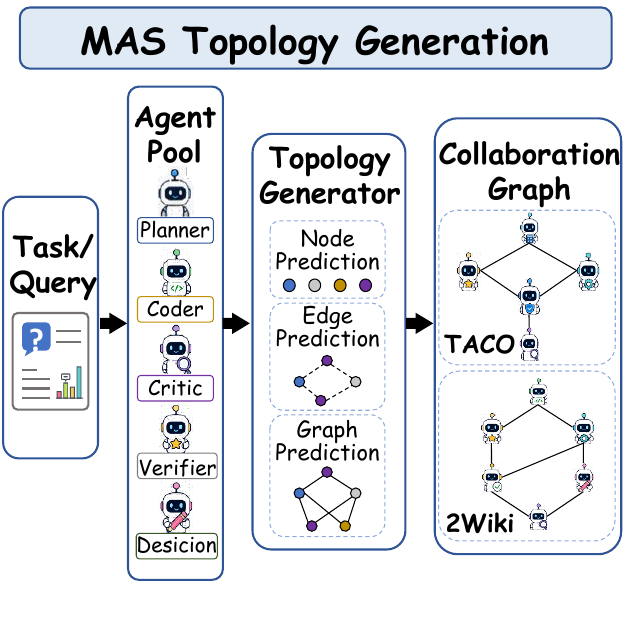}
        \caption{Task Illustration}
        \label{fig:mas-topology-generation}
    \end{subfigure}
    \hfill
    \begin{subfigure}[t]{0.24\linewidth}
        \centering
        \includegraphics[width=\linewidth]{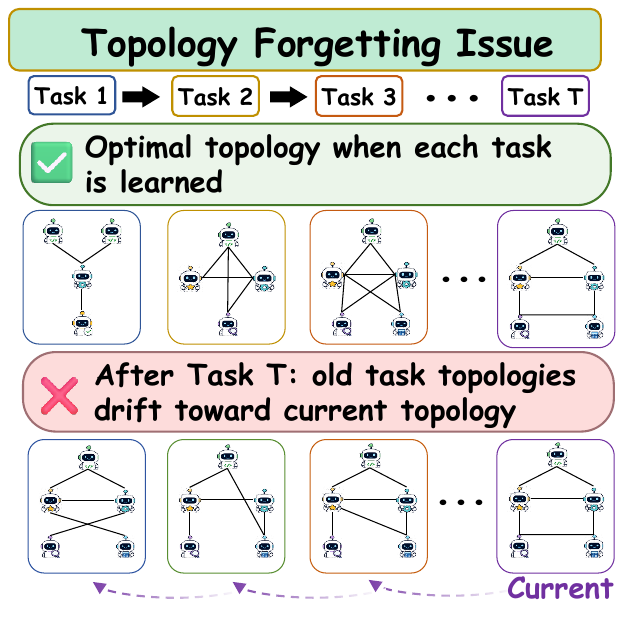}
        \caption{Topology Forgetting}
        \label{fig:topology-forgetting-issue}
    \end{subfigure}
    \hfill
    \begin{subfigure}[t]{0.24\linewidth}
        \centering
        \includegraphics[width=\linewidth]{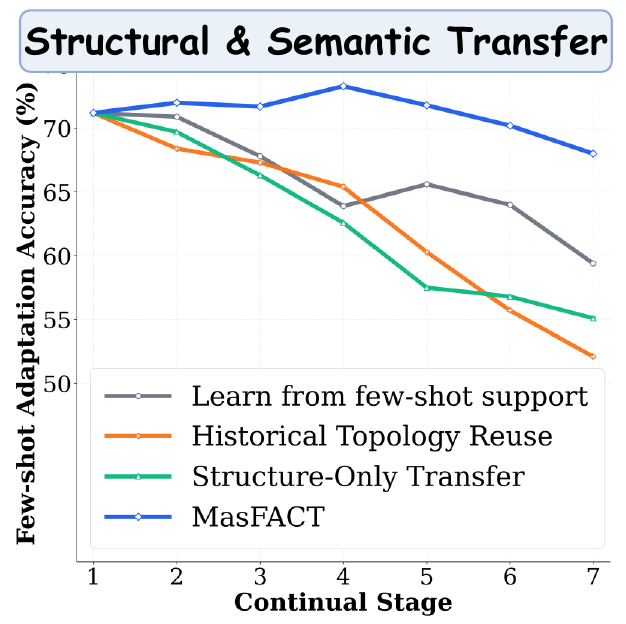}
        \caption{Challenge 1}
        \label{fig:challenge-1}
    \end{subfigure}
    \hfill
    \begin{subfigure}[t]{0.24\linewidth}
        \centering
        \includegraphics[width=\linewidth]{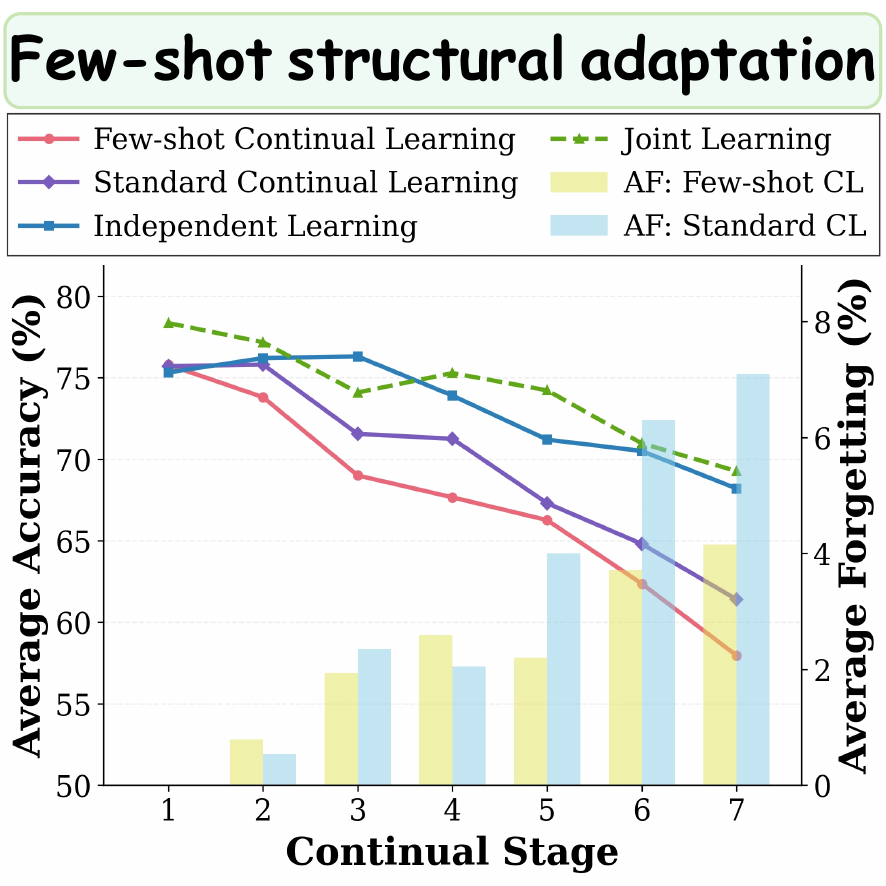}
        \caption{Challenge 2}
        \label{fig:challenge-2}
    \end{subfigure}

    \caption{Overview of MAS topology and related challenges. (a) An illustration of MAS topology generation. (b) Optimization of current task leads to topology interference on previous tasks. (c-d) Two challenges of continual MAS topology learning task.}
    \label{fig:mas-four-panels}
\end{figure}

Recent studies have explored automated MAS topology generation through search-based optimization and generative graph modeling~\cite{leong2025amas}, but largely assumed static task distributions or isolated-task optimization~\cite{chen2023agentverse}. 
In realistic deployments, MAS must continually adapt to evolving task streams, where previously effective collaboration structures should be retained and reused rather than repeatedly rediscovered or overwritten.
As illustrated in Figure~\ref{fig:topology-forgetting-issue}, we call this failure mode \emph{topology forgetting}: the topology generator gradually shifts away from the communication structures required by previous tasks when it adapts to new tasks, losing the ability to reconstruct them. 
Unlike standard parameter forgetting, topology forgetting stems from cross-task misalignment in topology space, where both relational structures and agent semantics may shift.



Therefore, we study continual MAS topology learning, where a topology generator must adapt to new tasks while preserving reusable collaboration topologies from prior tasks. 
Historical communication structures therefore should be treated not as isolated optimization outcomes, but as reusable structural knowledge transferable across tasks. 
Existing topology optimization frameworks are not designed to address this setting: search-based methods overwrite past structures during optimization~\cite{li2025adaptive,wang2025agentdropout}, while generative topology models suffer from distribution drift without explicit structural memory~\cite{leong2025amas,fan2026todycomm}. As a result, current approaches typically trade the short-term adaptation at the cost of degenerating long-horizon knowledge. This problem thus introduces two key challenges:

\textbf{Challenge \MakeUppercase{\romannumeral 1}: Joint structural-semantic transfer.}
MAS topologies encode both communication links and agent-level semantics (\eg, roles, tools, latent reasoning, and activation behaviors), requiring alignment across both relational structure and functional semantics. As shown in Figure~\ref{fig:mas-four-panels}(c), transferring structure alone yields suboptimal performance.

\textbf{Challenge \MakeUppercase{\romannumeral 2}: Few-shot adaptation under structural interference.}
Sparse feedback signals make na\"ive updates prone to overfitting and forgetting. \nop{This raises a topology-level trade-off between stability and plasticity, where continual adaptation must balance the reuse of historical collaboration patterns with the discovery of new task-specific structures.}
As shown in Figure~\ref{fig:mas-four-panels}(d), few-shot continual learning induces sharp topology forgetting from early stages, requiring a stability--plasticity balance to prevent the generator from drifting away from high-utility topology distributions.

To address these challenges, we propose \model (\textbf{F}actorized \textbf{A}ligned \textbf{C}onsensus \textbf{T}opology transfer), a geometry-aware prior-to-posterior framework for continual MAS topology learning. \model maintains a memory of factorized topology priors, retrieves and aligns the most transferable prior to each new task via Fused Gromov--Wasserstein optimal transport~\cite{titouan2019optimal}, and learns sparse task-specific posteriors around it. The factorized prior enables joint alignment of relational structure and agent semantics, while prior-centered posterior updates residual topology edits, balancing the task adaptation and knowledge retention under PAC-Bayes~\cite{mcallester1998some} complexity control.

We further introduce a continual MAS evaluation protocol spanning task-, domain-, and class-incremental regimes across four reasoning domains. The protocol organizes 17 reasoning datasets into 12 continual scenarios, covering diverse topology requirements, such as verification, decomposition, execution-guided debugging, and evidence aggregation. Across all scenarios, integrating \model with three representative topology generators improves average accuracy (AA) by 9.3\% and reduces average forgetting (AF) by 69.4\%, demonstrating robust gains over single MAS baselines.



In summary, our contributions are three-fold:

\begin{itemize}[leftmargin=*]
     \item {\bf Problem formulation:} To our best knowledge, we are the first to formulate continual MAS topology learning and identify topology forgetting as a distinct topology-level failure mode.
    \item {\bf Novel framework:} 
    We propose \model, a geometry-aware posterior transfer framework that aligns and reuses historical high-utility topologies, while constraining new-task adaptation through conservative posterior transfer.
    \item {\bf Evaluation:} We design a hierarchical continual MAS evaluation protocol covering task-, domain-, and class-level task evolution and show consistent improvements on both accuracy and forgetting metrics across diverse settings.
    
\end{itemize} 
\vspace{-1.5mm}

%% file: text/03-definition.tex
\section{Problem Definition}
\label{definition}

\paragraph{Attributed MAS topology.}
We formalize a MAS as a topology-controlled computational graph. Each agent is an \textit{attributed computational operator} whose behavior is governed by multi-dimensional states, such as temporal roles, decision status, and operational memory.
For task $\gT_t$, its topology is an attributed directed graph $\gG_t=(\mB_t,\mX_t,\vmu_t)$, where $\mB_t$ is a feasible discrete communication topology, $\mX_t$ encodes agent attributes, and $\vmu_t$ describes task-specific agent importance. This representation separates communication structure from agent labels, allowing agents with different names to play similar structural-functional roles and identically named agents to behave differently across tasks.

\paragraph{Continual MAS topology learning.}
We consider a task stream $\{\gT_0,\gT_1,\dots,\gT_K\}$, where a data-rich base task $\gT_0$ initializes the topology generator and each subsequent task $\gT_t$ provides only a few-shot support set $\gS_t$. For each task, the system must construct a valid topology $\mB_t$. Under the continual constraint, historical data are largely inaccessible, and old-task evaluation permits read-only inference without gradient updates or test-time adaptation. The central challenge is to adapt rapidly to $\gS_t$ while constraining topology plasticity, especially when agent operator spaces are unaligned across tasks.

\paragraph{Learning objective.}
At continual stage $t$, the topology generator outputs a topology distribution $\pi_t(\cdot\mid\gS_t)$ over the feasible topology space $\gB$. For a sampled topology $\mB\sim\pi_t(\cdot\mid\gS_t)$, the MAS executes the task under $\mB$ and receives an empirical utility $\widehat U_{\gS_t}(\mB)$. Equivalently, we define the empirical topology risk as
\begin{equation}
\widehat{\gR}_{\gS_t}(\pi_t)
=
\E_{\mB\sim\pi_t(\cdot\mid\gS_t)}
\left[\widehat{\ell}_{\gS_t}(\mB)\right].
\label{eq:empirical-topology-risk}
\end{equation}
Let $\gR_i(\pi_t)$ denote the population topology risk of applying the current topology distribution to task $\gT_i$. Ideally, after observing tasks up to stage $t$, the learner would minimize the aggregate risk $\min_{\pi_t}\frac{1}{t+1}\sum_{i=0}^{t}\gR_i(\pi_t)$. This objective is infeasible because historical data are not fully replayable, new tasks provide only few-shot feedback, agent spaces may be unaligned, and na\"ive optimization can overwrite previously effective structural patterns. We therefore use historical topology knowledge to guide current-task adaptation while preventing topology forgetting.

%% file: text/04_method_v2.tex
\section{Methodology}
\label{sec:method}

\begin{figure*}[t]
    \centering
    \includegraphics[width=0.98\textwidth]{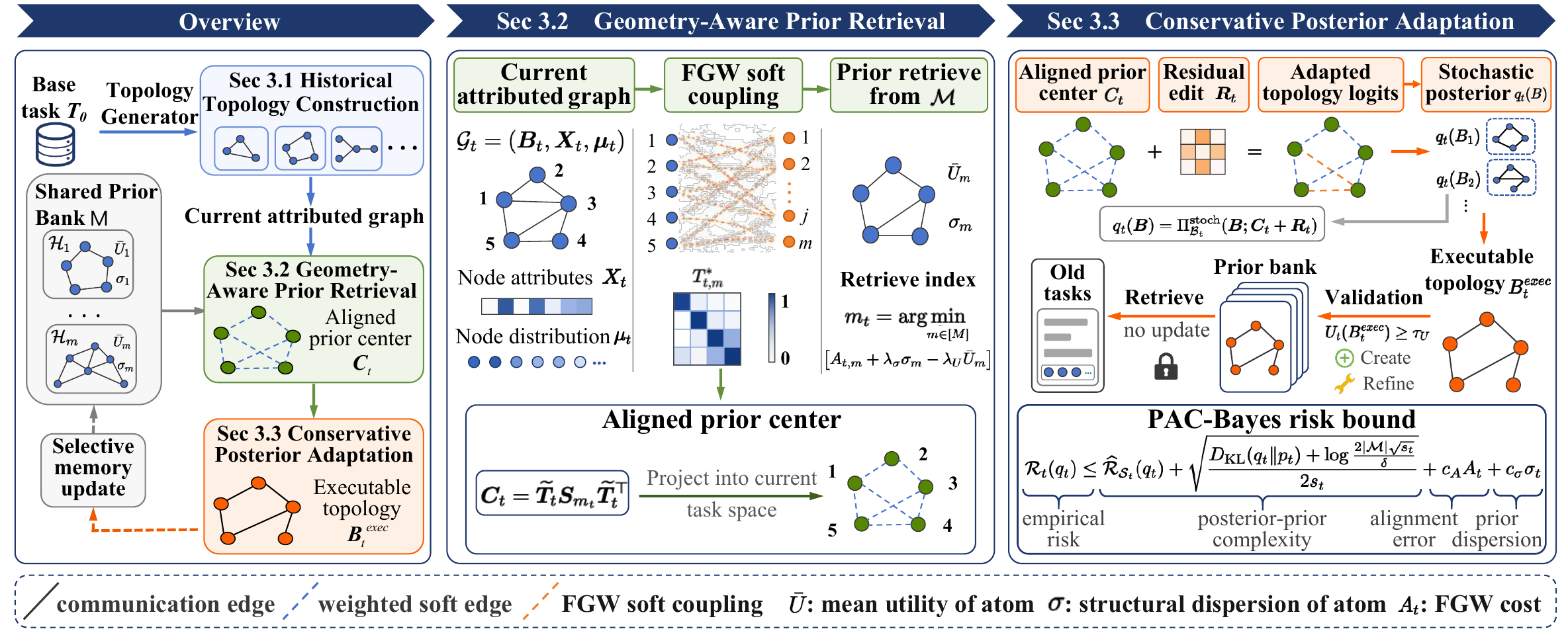}
    \caption{
    Overview of \model.
    The framework has three stages:
    historical topology construction (Sec.~\ref{sec:historical-topology}),
    geometry-aware prior retrieval (Sec.~\ref{sec:prior-retrieval}),
    and conservative posterior adaptation (Sec.~\ref{sec:posterior-adaptation}).
    }
    \label{fig:fact-fgw-overview}
\end{figure*}

Building on the continual MAS topology learning objective in Eq.~\eqref{eq:empirical-topology-risk}, we propose \model, a geometry-aware posterior transfer framework for mitigating topology forgetting under few-shot task evolution. 
As shown in Figure~\ref{fig:fact-fgw-overview}, \model converts high-utility historical topologies into transferable structural priors, aligns them to the current agent space, and performs sparse posterior adaptation under PAC-Bayes complexity control.

Specifically, \model transfers historical structural knowledge through a three-stage pipeline:
\begin{itemize}[leftmargin=*]
    \item \textbf{Factorized Historical Topology Construction}: \model consolidates high-utility topologies into prior atoms that encode reusable collaboration structures beyond task-specific agent identities.
    \item \textbf{Geometry-Aware Prior Retrieval}: Given a new task, \model retrieves and aligns the most transferable prior by comparing structural relations and agent attributes.
    \item \textbf{Conservative Posterior Adaptation}: \model learns a sparse topology posterior around the aligned prior, enabling adaptation while limiting structural drift.
\end{itemize}
\vspace{-1.5mm}

The complete training and inference procedure of \model is provided in Appendix~\ref{app:algorithm}.

\subsection{Factorized Historical Topology Construction}
\label{sec:historical-topology}

We represent historical structural knowledge as a bank of factorized prior atoms $\gM=\{\gH_m\}_{m=1}^{M}$, where each atom $\gH_m=(\mS_m,\mX_m,\bm{\nu}_m)$ contains three FGW-transportable components:

\begin{itemize}[leftmargin=*]
    \item Consensus topology $\mS_m$: a directed topology encoding reusable communication structure.
    \item Prototype attributes $\mX_m$: prototype agent attributes capturing transferable functional semantics.
    \item Node measure $\vnu_m$: a node measure over prototype nodes for optimal-transport alignment.
\end{itemize}
This factorization enables topology priors to be transferred across tasks with different agent identities, role semantics, and graph cardinalities. Each atom further maintains historical utility $\bar U_m$ and structural dispersion $\sigma_m$ as retrieval metadata, which are kept outside $\gH_m$ to separate retrieval criteria from the FGW-transportable graph components that define the alignment geometry.

\subsection{Geometry-Aware Retrieval via Optimal Transport}
\label{sec:prior-retrieval}

Given a current task $\gT_t$, we construct a task-conditioned scaffold graph 
$\gG_t^0=(\mB_t^0,\mX_t,\vmu_t)$, where $\mB_t^0$ is produced by the topology generator before posterior adaptation and serves only as a structural proxy for alignment. 
Let $\gH_m=(\mS_m,\mX_m,\vnu_m)$ be a historical prior atom. 
Since $\gG_t^0$ and $\gH_m$ may differ in node cardinality, agent semantics, and node ordering, index-based retrieval cannot reliably determine whether a historical topology is transferable. 
We therefore formulate prior retrieval as a geometry-aware optimal transport problem.

\paragraph{FGW-based prior retrieval.}
We align current agents to prototype nodes through a soft coupling
\begin{equation}
\mT\in\gU(\vmu_t,\bm{\nu}_m)
:=
\left\{
\mT\in\R_+^{N_t\times N_m}
~\middle|~
\mT\vone_{N_m}=\vmu_t,\;
\mT^\top\vone_{N_t}=\bm{\nu}_m
\right\},
\label{eq:transport-polytope}
\end{equation}
where $N_t$ and $N_m$ are the numbers of current agents and prototype nodes. 
The coupling $\mT$ defines a probabilistic correspondence between the current task graph and a historical prior atom. 

However, matching node attributes alone is insufficient, as topology utility is fundamentally determined by its \emph{relational structure}. We therefore adopt the \emph{Fused Gromov-Wasserstein} distance to jointly align agent attributes and graph structures, obtaining the optimal coupling by
\begin{equation}
\mT_{t,m}^*
=
\argmin_{\mT\in\gU(\vmu_t,\vnu_m)}
D_{\mathrm{FGW}}(\gG_t^0,\gH_m;\mT),
\label{eq:fgw-coupling}
\end{equation}
where the FGW objective is
\begin{equation}
D_{\mathrm{FGW}}(\gG_t^0,\gH_m;\mT)
=
(1-\rho)\langle \mC^X_{t,m},\mT\rangle
+
\rho\langle \tC^B_{t,m},\mT\otimes\mT\rangle
+
\varepsilon\Omega(\mT).
\label{eq:fgw-objective}
\end{equation}
Here, $\rho$ controls the semantic--structural trade-off. The node-attribute cost matrix $\mC^X_{t,m}$ is computed from current agent attributes $\mX_t$ and prototype attributes $\mX_m$, with $[\mC^X_{t,m}]_{ij}=c_X([\mX_t]_i,[\mX_m]_j)$. It measures the functional mismatch between current agent $i$ and prototype node $j$. Assigning large transport mass $\mT_{ij}$ to semantically incompatible nodes increases the alignment cost.

The relational cost tensor $\tC^B_{t,m}$ is computed from $\mB_t^0$ and $\mS_m$, with
$[\tC^B_{t,m}]_{(i,i'),(j,j')}=c_B([\mB_t^0]_{ii'},[\mS_m]_{jj'})$, comparing current directed relation $(i,i')$ with prototype relation $(j,j')$. 
The product coupling $\mT\otimes\mT$, defined by
$[\mT\otimes\mT]_{(i,i'),(j,j')}=[\mT]_{ij}[\mT]_{i'j'}$, lifts node-level matching to relation-level matching. 
Thus, FGW penalizes transport plans that align semantically incompatible agents or fail to preserve directed communication patterns. The entropy term $\Omega(\mT)$ stabilizes the soft coupling; additional details are provided in Appendix~\ref{app:fgw-details}.

The optimal coupling provides both a soft agent-to-prototype correspondence and an alignment cost
\begin{equation}
A_{t,m}
=
D_{\mathrm{FGW}}(\gG_t^0,\gH_m;\mT_{t,m}^*),
\label{eq:alignment-cost}
\end{equation}
which serves as the transferability score of atom $\gH_m$ for task $\gT_t$. We retrieve the atom that balances geometric transferability, structural concentration, and historical utility:
\begin{equation}
m_t
=
\argmin_{m\in[M]}
\left[
A_{t,m}
+
\lambda_{\sigma}\sigma_m
-
\lambda_U\bar U_m
\right].
\label{eq:retrieval-rule}
\end{equation}
After retrieval, we abbreviate the selected coupling as
$\mT_t^*:=\mT_{t,m_t}^*$, with $A_t=A_{t,m_t}$ and $\sigma_t=\sigma_{m_t}$.

\nop{To push the selected consensus topology into the current agent space, we row-normalize the selected coupling:
\begin{equation}
\widetilde{\mT}_t
=
\operatorname{diag}(\vmu_t)^{-1}\mT_t^*.
\label{eq:normalized-coupling}
\end{equation}
The normalized coupling $\widetilde{\mT}_t$ maps each current agent to a distribution over prototype nodes. The aligned prior center is then
\begin{equation}
\mC_t
=
\widetilde{\mT}_t
\mS_{m_t}
\widetilde{\mT}_t^\top.
\label{eq:aligned-prior-center}
\end{equation}
Thus, $\mC_t$ is a continuous structural prior over the current agent set, rather than a sampled execution graph. It provides the geometry-aligned reference around which the posterior topology distribution is conservatively adapted.}

\paragraph{Task-space prior projection.}
To transfer the selected consensus topology into the current agent space, we row-normalize the selected coupling and project the atom topology:
\begin{equation}
\widetilde{\mT}_t
=
\operatorname{diag}(\vmu_t)^{-1}\mT_t^*,
\qquad
\mC_t
=
\widetilde{\mT}_t
\mS_{m_t}
\widetilde{\mT}_t^\top .
\label{eq:aligned-prior-center}
\end{equation}
The normalized coupling $\widetilde{\mT}_t$ maps each current agent to a distribution over prototype nodes, and $\mC_t$ is the resulting continuous structural prior over the current agent set. It provides the geometry-aligned reference around which the posterior topology distribution is conservatively adapted.

\subsection{PAC-Bayes-Guided Conservative Posterior Adaptation}
\label{sec:posterior-adaptation}

After retrieval and alignment, the aligned center $\mC_t$ defines a task-specific structural prior for the current agent space. \model adapts to the new task by learning a stochastic posterior around this prior, rather than freely searching the entire topology space. Specifically, we define the pushed-forward topology prior and the residual posterior as
\begin{equation}
p_t(\mB)
=
\Pi_{\gB_t}^{\mathrm{stoch}}(\mB;\mC_t),
\qquad
q_t(\mB)
=
\Pi_{\gB_t}^{\mathrm{stoch}}(\mB;\mC_t+\mR_t),
\label{eq:posterior-policy}
\end{equation}
where $\mB\in\gB$ is a feasible communication topology, $p_t$ is the FGW-pushed prior, $q_t$ is the residual posterior, and $\mR_t$ is a learnable residual score matrix for task-specific structural edits. The operator $\Pi_{\gB_t}^{\mathrm{stoch}}$ is a masked stochastic mapping from topology scores to valid graphs, implemented with a straight-through Gumbel-Sigmoid estimator for differentiable sampling. The feasible mask prevents invalid edges from being sampled, while the sparsity regularization on $\mR_t$ encourages the posterior to modify only necessary edges around the FGW-aligned prior. See Appendix~\ref{app:posterior-details} for $\Pi_{\gB_t}^{\mathrm{stoch}}$.

Training optimizes the expected task risk under $q_t$. During evaluation, \model executes a high-probability topology from $q_t$ without gradient updates or memory writing.

\paragraph{Theoretical guarantee.} Let $\gR_t(q_t)$ and $\widehat{\gR}_{\gS_t}(q_t)$ denote the true and empirical topology risks defined in Section~\ref{definition}. We assume a \emph{local FGW stability condition}: small perturbations in FGW geometry around the aligned prior induce bounded changes in expected topology risk. Under this assumption, the transfer distortion is controlled by the alignment cost $A_t$, while the uncertainty introduced by compressing historical topologies into a shared atom is captured by $\sigma_t$.

\begin{theorem}[Geometry-Aware PAC-Bayes Transfer Bound]
\label{thm:pac-bayes-transfer}
Assume the topology-induced loss is bounded in $[0,1]$. For support set $\gS_t$ of size $s_t$, with probability at least $1-\delta$ over the draw of $\gS_t$, for any topology posterior $q_t$,
\begin{equation}
\gR_t(q_t)
\le
\widehat{\gR}_{\gS_t}(q_t)
+
\sqrt{
\frac{
\KL(q_t\Vert p_t)
+
\log\frac{2|\gM|\sqrt{s_t}}{\delta}
}{2s_t}}
+
c_A A_t
+
c_\sigma \sigma_t .
\label{eq:pac-bayes-transfer-bound}
\end{equation}
Here, $\KL(q_t\Vert p_t)$ is the posterior-prior complexity relative to the FGW-pushed prior, $|\gM|$ accounts for finite prior-bank selection, $A_t$ is the FGW alignment error, and $\sigma_t$ is the compression uncertainty of the selected prior. The constants $c_A$ and $c_\sigma$ quantify risk sensitivity to alignment distortion and prior compression, and are not optimization hyperparameters.
\end{theorem}

\paragraph{Remarks.} Full risk definitions, theoretical assumptions, proof details, and the derivation of the bound-induced training surrogate are provided in Appendix~\ref{app:posterior-details}--~\ref{app:prior-bank-update}. The bound motivates minimizing an empirical surrogate that balances current-task fit and structural drift from the aligned prior:
\begin{equation}
\Ls_{\mathrm{train}}
=
\widehat{\gR}_{\gS_t}(q_t)
+
\lambda_{\mathrm{KL}}\KL(q_t\Vert p_t)
+
\lambda_R\|\mR_t\|_1.
\label{eq:train-loss}
\end{equation}
The KL term keeps the posterior close to the FGW-aligned prior, while the sparse residual penalty restricts adaptation to local topology corrections. Together with the retrieval rule in Eq.~\ref{eq:retrieval-rule}, this converts few-shot topology adaptation into a prior-centered posterior transfer problem, coupling current-task plasticity with protection against topology forgetting.

%% file: text/05-experiments.tex
\section{Experiments}
\label{experiment}

\subsection{Experimental setup}
\noindent\textbf{Hierarchy-aware continual MAS evaluation protocol.}
We instantiate the protocol along three axes of MAS evolution: task-level reasoning-paradigm shifts, domain-level difficulty and evaluation-style shifts, and class-level subskill evolution with coarse/fine splits for short-horizon adaptation and long-horizon stability. Across four reasoning families with distinct collaboration pressures, we organize 17 public datasets into 12 continual scenarios for evaluating performance, forgetting, and adaptation robustness; details are in Appendix~\ref{app:benchmark-details}.

\noindent\textbf{Baselines.}
We compare \model with four groups of baselines: single-agent methods (Vanilla, CoT, and Self-Consistency CoT~\cite{wang2022self}); fixed MAS topologies (Chain, Tree, and Complete Graph~\cite{qian2024scaling}); adaptive topology methods (GDesigner~\cite{zhang2024g}, AgentDropout~\cite{wang2025agentdropout}, ARGDesigner~\cite{li2026assemble}, AGP~\cite{li2025adaptive}, and MasRouter~\cite{yue2025masrouter}); and continual MAS variants with replay~\cite{rolnick2019experience}. 
We also evaluate the plug-and-play capability of \model across different MAS architectures. 
Details are provided in Appendix~\ref{app:baseline-details}.

\noindent\textbf{Implementation and Evaluation.}
All experiments use a unified graph-based MAS backbone following GDesigner, with fixed agent pools within each task family to isolate the effect of topology learning. 
We use Llama-3.1-8B-Instruct\footnote{We use Llama-3.1-8B-Instruct as a widely adopted open-source backbone with a practical performance--efficiency--cost trade-off. 
\model is backbone-agnostic and can use other LLMs without changing the continual topology learning procedure.}, 200 base-stage samples, and 10 support samples per continual stage. 
We report Average Accuracy (AA) and Average Forgetting (AF), where lower AF indicates less forgetting and negative AF denotes positive backward transfer.

\subsection{Main Results}
\input{table/Table1}
Table~\ref{tab:main-class-results} reports class-level continual learning results across four task families. \model achieves the strongest overall performance, improving average AA while reducing average AF below zero. Unlike existing topology generators and routing-based methods, which optimize each new stage around task-specific topologies and therefore retain positive AF, \model converts continual adaptation into backward transfer on average, showing that previously learned collaboration patterns are not merely protected but can also benefit later evaluation. \model addresses this limitation by reusing factorized consensus priors, aligning them to the current operator graph with FGW, and restricting adaptation to sparse residual edits around the aligned prior center. These results validate continual topology preservation as the main source of \model's advantage.

\input{table/Table2}
\noindent\textbf{Broader-Stream Continual Learning.}
Table~\ref{tab:domain-task-results} evaluates \model on broader streams combining domain shifts and task transitions. \model achieves the best overall AA--AF trade-off, indicating that its topology memory is not tied to a single benchmark decomposition. Although several adaptive routing or graph baselines remain competitive in AA, they still show positive forgetting after sequential adaptation. In contrast, \model attains negative average AF, suggesting that transferred topology priors can improve later evaluation on previous tasks rather than merely slow degradation.

\input{table/Table3}
Table~\ref{tab:finegrained-results} evaluates a more demanding setting where MMLU-Pro, MATH, and TACO are decomposed into 13, 7, and 8 semantically close increments. 
\model is the only method that consistently obtains negative AF, showing that its topology priors not only mitigate forgetting but can also induce positive backward transfer when later tasks refine useful collaboration patterns. The table also exposes a \emph{MAS-specific failure mode}: replay is not universally beneficial for topology generators. It can also reduce AA or increase forgetting, as seen for GDesigner on MMLU-Pro and ARGDesigner on MATH, indicating that forgetting arises not only from sample distribution shift, but also from drift in the structure-generation policy. 

Replay without topology constraints may mix incompatible collaboration modes and blur fine-grained specialization. This effect becomes more evident in highly similar task increments. Thus, task similarity does not necessarily imply topology compatibility. Over-replay can dilute current topology and harm specialization. 
\model avoids this by factorizing historical structures into consensus priors and task-specific residuals, then applying FGW-constrained residual adaptation.

\input{table/Table4}
\noindent\textbf{Plug-and-Play Continual Topology Transfer.}
To evaluate whether \model is tied to a specific topology generator, we attach its continual topology layer to three representative recent and widely adopted paradigms of adaptive MAS optimizers, AGP~\cite{li2025adaptive}, GDesigner~\cite{zhang2024g}, and ARGDesigner~\cite{li2026assemble}, covering pruning-based, graph-generation-based, and autoregressive graph-design paradigms. 
Implementation details are provided in Appendix~\ref{app:plugin-baselines}. 
As shown in Table~\ref{tab:cross-mas-plugin}, \model consistently enhances performance across all backbones in class-, domain-, and task-incremental continual learning settings. 
This supports its role as an architecture-agnostic continual topology layer rather than a generator-specific trick. 
The gains indicate that structural priors provide not only regularization, but also better topology initialization and a more stable search direction, especially for dynamic optimizers that otherwise adapt aggressively and drift from previously useful collaboration patterns.

\input{table/Table5}
\noindent\textbf{Ablation Study.}
Table~\ref{tab:ablation} examines the individual contribution of each core design in \model.
\emph{(1) Historical prior.} Removing the historical topology prior causes the largest degradation, showing that continual MAS topology learning cannot be solved by current-task adaptation alone; reusable collaboration structures must be explicitly retained. 
\emph{(2) Geometry-aware alignment.} The alignment variants confirm that cross-task transfer requires jointly matching both topology geometry and agent semantics. Euclidean alignment performs worse than the no-FGW variant, suggesting that index-wise correspondence can misalign agents under role or node-order drift. GW-only alignment remains inferior to full FGW, because relational structure alone cannot disambiguate agent semantics. Full FGW provides a more reliable transfer map by jointly aligning agent attributes and communication relations. 
\emph{(3) Conservative posterior adaptation.} The residual and complexity-control ablations reveal \model's stability--plasticity mechanism. Removing residual editing weakens task-specific adaptation, whereas removing the PAC-Bayes constraint or replacing it with an L2 penalty increases forgetting. This confirms that \model's gains come from prior-centered posterior control rather than generic regularization.

\begin{figure}[t]
    \centering
    \begin{subfigure}[t]{0.58\linewidth}
        \centering
        \includegraphics[width=\linewidth]{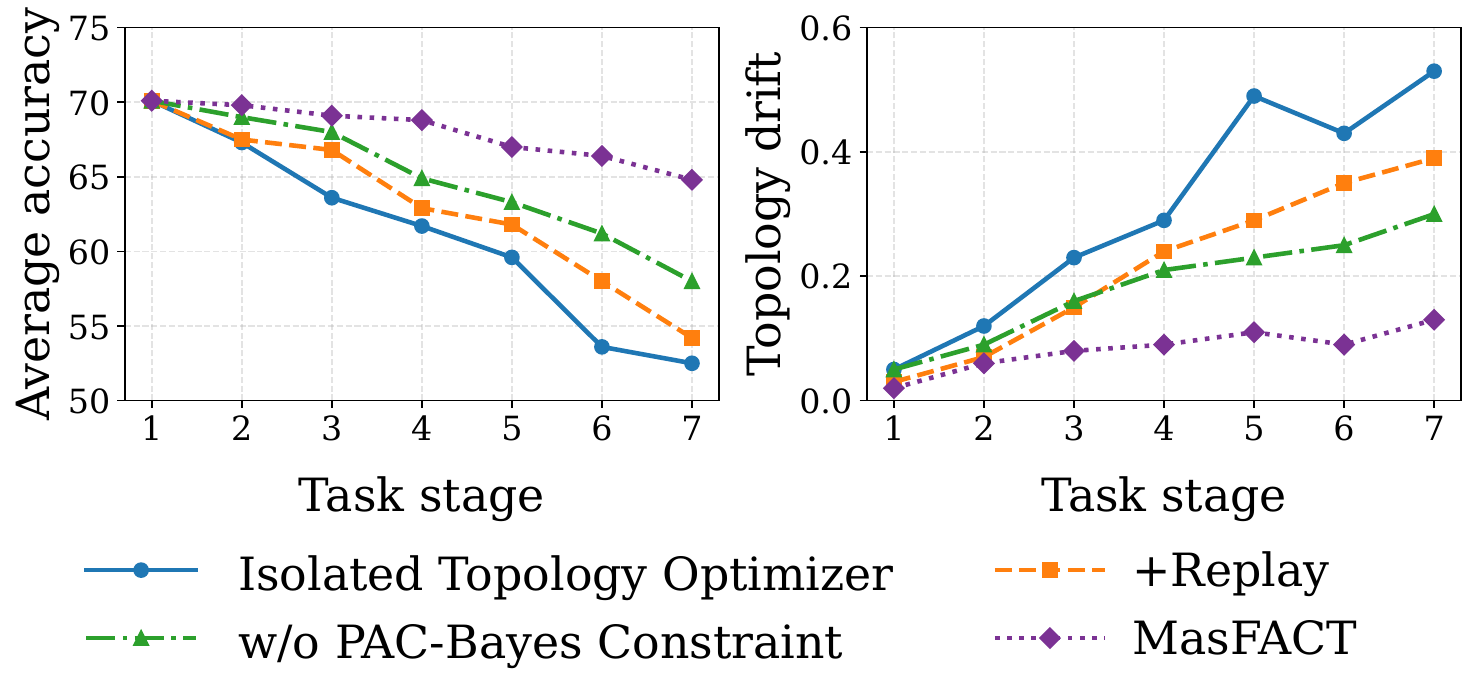}
        \caption{Topology forgetting dynamics.}
        \label{fig:topology-forgetting-dynamics}
    \end{subfigure}
    \hfill
    \begin{subfigure}[t]{0.41\linewidth}
        \centering
        \includegraphics[width=\linewidth]{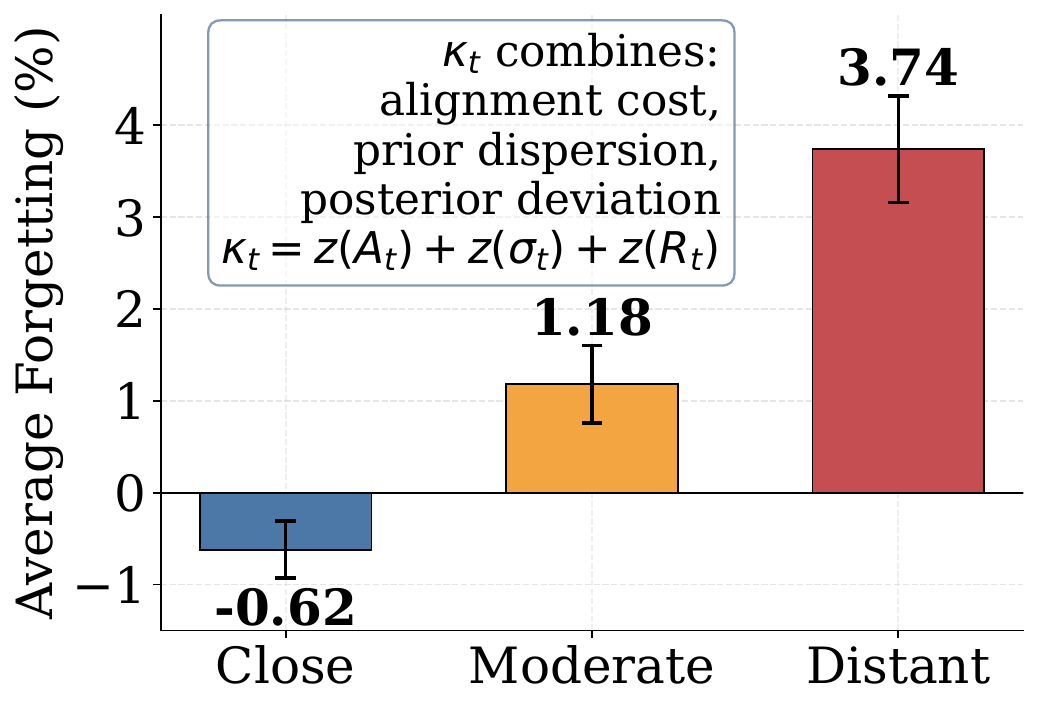}
        \caption{Forgetting by prior-transfer distance.}
        \label{fig:transfer-complexity}
    \end{subfigure}
    \caption{
    Mechanistic analysis of structural forgetting.
    Left: topology drift tracks old-task accuracy degradation.
    Right: forgetting increases as the retrieved prior becomes more distant.
    }
    \label{fig:dynamics-and-complexity}
\end{figure}

\noindent\textbf{Mechanistic Analysis of Structural Forgetting.}
We examine whether continual MAS degradation is coupled with drift in the learned communication topology. 
As shown in Figure~\ref{fig:dynamics-and-complexity}(a), the isolated topology optimizer suffers a sharp old-task accuracy drop as topology drift increases, indicating that new-task adaptation moves the generator away from previously effective collaboration patterns. 
Replay reduces but does not eliminate this drift, while removing the PAC-Bayes constraint still allows posterior updates to accumulate structural deviation despite aligned priors. 
In contrast, \model maintains the flattest accuracy trajectory and the lowest topology drift, supporting that continual MAS forgetting is a topology-level phenomenon rather than merely a sample-distribution issue.

Figure~\ref{fig:dynamics-and-complexity}(b) links this trend to the PAC-Bayes transfer analysis. 
We group stages into close, moderate, and distant prior-transfer regimes by $\kappa_t$, which combines FGW alignment cost, prior dispersion, and posterior deviation. 
Forgetting grows monotonically with prior-transfer distance: close transfers can yield backward transfer, whereas distant transfers cause substantial forgetting. 
The monotonic pattern supports our transfer-complexity term, showing that retention benefits only from well-aligned, stable priors rather than indiscriminate prior reuse. 
This explains why \model must jointly control alignment mismatch, prior instability, and posterior drift.
Sensitivity results in Appendix~\ref{app:hyperparameter} confirm that this trend is not due to hyperparameter tuning.

%% file: table/Table1.tex
\begin{table*}[htb]
\centering
\footnotesize
\setlength{\tabcolsep}{3.0pt}
\renewcommand{\arraystretch}{1.12}
\caption{
Class-level continual learning results. 
$\dagger$ denotes replay-based variants. Results are averaged over five random seeds. 
Results with standard deviations are reported in Appendix~\ref{app:additional-experiments}.
}
\label{tab:main-class-results}

\resizebox{\textwidth}{!}{%
\begin{tabular}{lcccccccccccccccccc}
\toprule
\multirow{3}{*}{\textbf{Method}}
& \multicolumn{4}{c}{MMLU-Pro}
& \multicolumn{4}{c}{\textbf{MATH}}
& \multicolumn{4}{c}{\textbf{TACO}}
& \multicolumn{4}{c}{\textbf{2Wiki}}
& \multicolumn{2}{c}{\textbf{Avg.}} \\
\cmidrule(lr){2-5}
\cmidrule(lr){6-9}
\cmidrule(lr){10-13}
\cmidrule(lr){14-17}
\cmidrule(lr){18-19}
& \multicolumn{2}{c}{\textbf{Few-shot}} & \multicolumn{2}{c}{\textbf{Standard}}
& \multicolumn{2}{c}{\textbf{Few-shot}} & \multicolumn{2}{c}{\textbf{Standard}}
& \multicolumn{2}{c}{\textbf{Few-shot}} & \multicolumn{2}{c}{\textbf{Standard}}
& \multicolumn{2}{c}{\textbf{Few-shot}} & \multicolumn{2}{c}{\textbf{Standard}}
& \multirow{2}{*}{\textbf{AA$\uparrow$}} & \multirow{2}{*}{\textbf{AF$\downarrow$}} \\
\cmidrule(lr){2-3}
\cmidrule(lr){4-5}
\cmidrule(lr){6-7}
\cmidrule(lr){8-9}
\cmidrule(lr){10-11}
\cmidrule(lr){12-13}
\cmidrule(lr){14-15}
\cmidrule(lr){16-17}
& \textbf{AA$\uparrow$} & \textbf{AF$\downarrow$} 
& \textbf{AA$\uparrow$} & \textbf{AF$\downarrow$} 
& \textbf{AA$\uparrow$} & \textbf{AF$\downarrow$} 
& \textbf{AA$\uparrow$} & \textbf{AF$\downarrow$} 
& \textbf{AA$\uparrow$} & \textbf{AF$\downarrow$} 
& \textbf{AA$\uparrow$} & \textbf{AF$\downarrow$} 
& \textbf{AA$\uparrow$} & \textbf{AF$\downarrow$} 
& \textbf{AA$\uparrow$} & \textbf{AF$\downarrow$} 
& & \\
\midrule

Vanilla
& 26.17 & -- 
& -- & -- 
& 35.79 & -- 
& -- & -- 
& 55.15 & -- 
& -- & -- 
& 39.26 & -- 
& -- & -- 
& 39.09 & -- \\

CoT-ICL
& 39.60 & 4.01
& -- & -- 
& 36.83 & 4.29
& -- & -- 
& 58.04 & 4.38
& -- & -- 
& 50.58 & 2.61
& -- & -- 
& 46.26 & 3.82 \\

SC(CoT)-ICL
& 40.52 & 4.95
& -- & -- 
& 37.88 & 6.16
& -- & -- 
& 59.22 & 3.03
& -- & -- 
& 58.36 & 2.15
& -- & -- 
& 49.00 & 4.07 \\

Chain
& 36.48 & -- 
& -- & -- 
& 36.92 & -- 
& -- & -- 
& 58.46 & -- 
& -- & -- 
& 50.62 & -- 
& -- & -- 
& 45.62 & -- \\

Tree
& 36.82 & -- 
& -- & -- 
& 36.73 & -- 
& -- & -- 
& 54.55 & -- 
& -- & -- 
& 52.74 & -- 
& -- & -- 
& 45.21 & -- \\

Complete
& 38.64 & -- 
& -- & -- 
& 37.63 & -- 
& -- & -- 
& 59.01 & -- 
& -- & -- 
& 54.68 & -- 
& -- & -- 
& 47.49 & -- \\

\midrule

AFLOW
& 40.12 & 3.97
& 42.37 & 4.32
& 42.18 & 3.37
& 48.85 & 3.92
& 62.84 & 3.68
& 64.12 & 4.21
& 60.48 & 2.90
& 61.97 & 3.45
& 52.87 & 3.73 \\

AFLOW$^\dagger$
& 42.52 & \underline{1.21}
& 42.89 & 2.19
& 44.39 & 2.03
& 49.63 & 1.75
& 64.75 & 2.64
& 65.48 & 2.13
& 63.06 & 2.16
& 64.73 & 2.29
& 54.68 & 2.05 \\

AgentDropout
& 41.07 & 3.13
& 42.48 & 3.83
& 43.54 & 4.18
& 49.88 & 4.96
& 62.72 & 3.17
& 63.14 & 3.53
& 60.07 & 1.81
& 62.84 & 1.97
& 53.22 & 3.32 \\

AgentDropout$^\dagger$
& 43.03 & 1.72
& 45.84 & 1.49
& 47.05 & \underline{1.56}
& 51.75 & 2.42
& 63.38 & \underline{1.70}
& 65.91 & 1.42
& 62.94 & 1.53
& 63.58 & \underline{1.10}
& 55.44 & \underline{1.62} \\

MasRouter
& 45.94 & 2.27
& 46.17 & 2.81
& 52.42 & 3.14
& 54.28 & 4.46
& 67.62 & 3.12
& 68.75 & 3.70
& 65.28 & 2.84
& 67.51 & 3.62
& 58.50 & 3.25 \\

MasRouter$^\dagger$
& \underline{47.40} & 1.43
& 47.53 & \underline{0.83}
& \underline{55.17} & 2.76
& \underline{58.56} & \underline{1.47}
& \underline{69.76} & 2.89
& 71.57 & 2.25
& \underline{69.16} & \underline{1.41}
& \underline{70.17} & 2.06
& \underline{61.31} & 1.89 \\

ARGDesigner
& 44.72 & 3.25
& 45.52 & 3.39
& 53.84 & 3.62
& 56.33 & 3.86
& 67.36 & 3.74
& 69.41 & 4.28
& 62.41 & 2.94
& 63.91 & 3.39
& 57.94 & 3.56 \\

ARGDesigner$^\dagger$
& 46.87 & 1.27
& \underline{48.69} & 1.33
& 54.21 & 1.88
& 57.83 & 1.59
& 68.82 & 2.21
& 70.24 & 1.92
& 65.36 & 1.65
& 67.51 & 2.08
& 59.94 & 1.74 \\

GDesigner
& 45.04 & 3.28
& 46.72 & 3.61
& 51.92 & 3.89
& 53.36 & 4.95
& 66.41 & 4.81
& 69.42 & 5.26
& 65.09 & 2.74
& 66.27 & 3.63
& 58.03 & 4.02 \\

GDesigner$^\dagger$
& 46.16 & 1.49
& 47.87 & 1.22
& 54.70 & 2.36
& 56.18 & 1.94
& 67.10 & 2.22
& \underline{72.35} & \underline{1.06}
& 67.79 & 1.84
& 69.84 & 2.17
& 60.25 & 1.79 \\

\rowcolor{gray!15}
\textbf{\model}
& \textbf{49.18} & \textbf{-0.19}
& \textbf{50.72} & \textbf{0.13}
& \textbf{56.07} & \textbf{-0.45}
& \textbf{61.68} & \textbf{0.21}
& \textbf{70.94} & \textbf{0.12}
& \textbf{73.80} & \textbf{-0.11}
& \textbf{69.92} & \textbf{-0.26}
& \textbf{71.74} & \textbf{0.08}
& \textbf{62.73} & \textbf{-0.06} \\

\midrule
Rel. Gains
& \textcolor{masfactgreen}{$\uparrow$3.76\%} & \textcolor{masfactgreen}{$\downarrow$115.70\%}
& \textcolor{masfactgreen}{$\uparrow$4.17\%} & \textcolor{masfactgreen}{$\downarrow$84.34\%}
& \textcolor{masfactgreen}{$\uparrow$1.63\%} & \textcolor{masfactgreen}{$\downarrow$128.85\%}
& \textcolor{masfactgreen}{$\uparrow$5.33\%} & \textcolor{masfactgreen}{$\downarrow$85.71\%}
& \textcolor{masfactgreen}{$\uparrow$1.69\%} & \textcolor{masfactgreen}{$\downarrow$92.94\%}
& \textcolor{masfactgreen}{$\uparrow$2.00\%} & \textcolor{masfactgreen}{$\downarrow$110.38\%}
& \textcolor{masfactgreen}{$\uparrow$1.10\%} & \textcolor{masfactgreen}{$\downarrow$118.44\%}
& \textcolor{masfactgreen}{$\uparrow$2.24\%} & \textcolor{masfactgreen}{$\downarrow$92.73\%}
& \textcolor{masfactgreen}{$\uparrow$2.32\%} & \textcolor{masfactgreen}{$\downarrow$103.70\%} \\

\bottomrule
\end{tabular}%
}
\vspace{2pt}
\end{table*}

%% file: table/Table2.tex
\begin{table*}[t]
\centering
\scriptsize
\setlength{\tabcolsep}{4.2pt}
\renewcommand{\arraystretch}{1.12}
\caption{
Domain- and task-level continual learning results.
$\dagger$ denotes replay-based variants. Results are averaged over five random seeds.
Results with standard deviations are reported in Appendix~\ref{app:additional-experiments}.
}
\label{tab:domain-task-results}
\resizebox{\textwidth}{!}{
\begin{tabular}{lcccccccccc}
\toprule
\multirow{3}{*}{\textbf{Method}}
& \multicolumn{8}{c}{\textbf{Domain-level continual learning}}
& \multicolumn{2}{c}{\textbf{Task-level continual learning}} \\
\cmidrule(lr){2-9}
\cmidrule(lr){10-11}
& \multicolumn{2}{c}{\textbf{Knowledge QA}}
& \multicolumn{2}{c}{\textbf{Math}}
& \multicolumn{2}{c}{\textbf{Code}}
& \multicolumn{2}{c}{\textbf{Multi-hop QA}}
& \multicolumn{2}{c}{\textbf{KQA$\rightarrow$Math$\rightarrow$Code$\rightarrow$MQA}} \\
\cmidrule(lr){2-3}
\cmidrule(lr){4-5}
\cmidrule(lr){6-7}
\cmidrule(lr){8-9}
\cmidrule(lr){10-11}
& \textbf{AA$\uparrow$} & \textbf{AF$\downarrow$} 
& \textbf{AA$\uparrow$} & \textbf{AF$\downarrow$} 
& \textbf{AA$\uparrow$} & \textbf{AF$\downarrow$} 
& \textbf{AA$\uparrow$} & \textbf{AF$\downarrow$} 
& \textbf{AA$\uparrow$} & \textbf{AF$\downarrow$} \\
\midrule

Vanilla
& 25.57 & -- 
& 29.28 & -- 
& 43.78 & -- 
& 35.85 & -- 
& 30.84 & -- \\

CoT-ICL
& 34.37 & 5.08
& 30.19 & 6.13
& 46.74 & 3.29
& 38.74 & 4.76
& 34.90 & 5.64 \\

SC(CoT)-ICL
& 32.54 & 7.46
& 33.82 & 6.24
& 50.46 & 3.40
& 41.51 & 3.24
& 37.47 & 5.63 \\

Chain
& 29.11 & -- 
& 30.63 & -- 
& 45.64 & -- 
& 32.03 & -- 
& 32.24 & -- \\

Tree
& 30.40 & -- 
& 34.29 & -- 
& 47.12 & -- 
& 37.81 & -- 
& 34.17 & -- \\

Complete Graph
& 32.24 & -- 
& 35.50 & -- 
& 41.71 & -- 
& 39.62 & -- 
& 36.06 & -- \\

\midrule

AFLOW
& 28.45 & 7.52
& 39.46 & 6.89
& 48.82 & 6.17
& 40.18 & 4.72
& 37.76 & 6.45 \\

AFLOW$^\dagger$
& 30.21 & 3.89
& 40.29 & 4.16
& 54.77 & 4.22
& 43.68 & 1.89
& 40.39 & \underline{4.12} \\

AgentDropout
& 33.15 & 5.83
& 39.59 & 5.62
& 48.91 & 4.03
& 42.94 & 5.19
& 36.72 & 8.12 \\

AgentDropout$^\dagger$
& 36.05 & 4.07
& 41.17 & 4.92
& 53.58 & \underline{2.28}
& 47.17 & 1.04
& 40.61 & 7.48 \\

GDesigner
& 45.62 & 6.33
& 42.93 & 5.14
& 54.46 & 4.89
& 44.92 & 5.23
& 41.81 & 6.95 \\

GDesigner$^\dagger$
& 47.73 & 4.20
& 45.52 & \underline{3.57}
& 57.13 & 5.04
& 47.97 & 1.26
& 44.21 & 5.15 \\

MasRouter
& 46.84 & 4.64
& 43.86 & 4.02
& 58.03 & 4.18
& 45.26 & 3.74
& 45.50 & 7.48 \\

MasRouter$^\dagger$
& \underline{49.16} & \underline{3.01}
& \underline{46.28} & 3.60
& \textbf{62.09} & 3.46
& 48.79 & \underline{0.72}
& \underline{49.55} & 5.02 \\

ARGDesigner
& 47.11 & 5.36
& 38.94 & 3.93
& 52.47 & 4.09
& 46.32 & 4.27
& 42.18 & 6.14 \\

ARGDesigner$^\dagger$
& 48.70 & 3.16
& 40.40 & 7.59
& 58.41 & 2.76
& \underline{49.32} & 1.90
& 49.13 & 4.83 \\

\rowcolor{gray!15}
\textbf{\model}
& \textbf{51.59} & \textbf{0.06}
& \textbf{47.91} & \textbf{0.57}
& \underline{61.14} & \textbf{-2.03}
& \textbf{52.31} & \textbf{-0.55}
& \textbf{51.67} & \textbf{2.07} \\

\midrule
Rel. Gains
& \textcolor{masfactgreen}{$\uparrow$4.94\%} & \textcolor{masfactgreen}{$\downarrow$98.01\%}
& \textcolor{masfactgreen}{$\uparrow$3.52\%} & \textcolor{masfactgreen}{$\downarrow$84.03\%}
& \textcolor{masfactgreen}{$\downarrow$1.53\%} & \textcolor{masfactgreen}{$\downarrow$189.04\%}
& \textcolor{masfactgreen}{$\uparrow$6.06\%} & \textcolor{masfactgreen}{$\downarrow$176.39\%}
& \textcolor{masfactgreen}{$\uparrow$4.28\%} & \textcolor{masfactgreen}{$\downarrow$49.76\%} \\

\bottomrule
\end{tabular}
}
\end{table*}

%% file: table/Table3.tex
\begin{table*}[t]
\centering
\small
\setlength{\tabcolsep}{5.2pt}
\renewcommand{\arraystretch}{1.08}
\caption{
Fine-grained class-level continual learning results.
$\dagger$ denotes replay-based variants.
Results are averaged over five random seeds. 
}
\label{tab:finegrained-results}
\resizebox{\textwidth}{!}{
\begin{tabular}{lcccccccc}
\toprule
\multirow{2}{*}{\textbf{Method}}
& \multicolumn{2}{c}{\textbf{MMLU-Pro (session=13)}}
& \multicolumn{2}{c}{\textbf{MATH (session=7)}}
& \multicolumn{2}{c}{\textbf{TACO (session=8)}}
& \multicolumn{2}{c}{\textbf{Average}} \\
\cmidrule(lr){2-3}
\cmidrule(lr){4-5}
\cmidrule(lr){6-7}
\cmidrule(lr){8-9}
& \textbf{AA}$\uparrow$ & \textbf{AF}$\downarrow$ 
& \textbf{AA}$\uparrow$ & \textbf{AF}$\downarrow$ 
& \textbf{AA}$\uparrow$ & \textbf{AF}$\downarrow$ 
& \textbf{AA}$\uparrow$ & \textbf{AF}$\downarrow$ \\
\midrule

AFLOW
& \score{29.77}{0.39} & \score{6.27}{0.11}
& \score{44.31}{0.12} & \score{4.32}{0.08}
& \score{47.92}{0.19} & \score{7.25}{0.04}
& \score{40.67}{0.40} & \score{5.95}{0.04} \\

AFLOW$^\dagger$
& \score{35.10}{0.46} & \score{2.78}{0.13}
& \score{47.76}{0.27} & \score{2.92}{0.04}
& \score{53.41}{0.21} & \score{4.36}{0.09}
& \score{45.42}{0.13} & \score{3.35}{0.08} \\

AgentDropout
& \score{32.67}{0.19} & \score{5.79}{0.04}
& \score{43.86}{0.33} & \score{3.52}{0.09}
& \score{48.21}{0.41} & \score{7.81}{0.11}
& \score{41.58}{0.28} & \score{5.71}{0.16} \\

AgentDropout$^\dagger$
& \score{39.33}{0.24} & \score{4.29}{0.12}
& \score{46.48}{0.09} & \score{3.31}{0.17}
& \score{54.17}{0.48} & \score{6.34}{0.25}
& \score{46.66}{0.17} & \score{4.65}{0.14} \\

GDesigner
& \score{41.67}{0.51} & \score{2.34}{0.19}
& \score{49.27}{0.24} & \score{6.25}{0.14}
& \score{65.29}{0.39} & \score{3.37}{0.26}
& \score{52.08}{0.31} & \score{3.99}{0.10} \\

GDesigner$^\dagger$
& \score{37.95}{0.64} & \score{2.93}{0.13}
& \score{52.39}{1.18} & \score{3.01}{0.14}
& \score{69.72}{0.56} & \underline{\score{2.89}{0.17}}
& \score{53.35}{0.84} & \score{2.94}{0.12} \\

MasRouter
& \score{40.10}{0.36} & \score{4.06}{0.26}
& \score{54.16}{0.37} & \score{5.31}{0.09}
& \score{63.57}{0.28} & \score{4.90}{0.11}
& \score{52.61}{0.16} & \score{4.76}{0.06} \\

MasRouter$^\dagger$
& \score{43.75}{0.20} & \score{1.95}{0.29}
& \underline{\score{58.36}{0.24}} & \underline{\score{1.19}{0.16}}
& \underline{\score{70.78}{0.19}} & \score{3.18}{0.09}
& \underline{\score{57.63}{0.53}} & \underline{\score{2.11}{0.09}} \\

ARGDesigner
& \score{37.56}{0.16} & \score{2.59}{0.35}
& \score{56.87}{0.04} & \score{2.79}{0.13}
& \score{65.67}{0.31} & \score{5.53}{0.14}
& \score{53.37}{0.09} & \score{3.64}{0.10} \\

ARGDesigner$^\dagger$
& \underline{\score{45.90}{0.17}} & \underline{\score{1.06}{0.11}}
& \score{54.50}{0.41} & \score{3.67}{0.09}
& \score{69.72}{0.26} & \score{3.83}{0.17}
& \score{56.71}{0.30} & \score{2.85}{0.03} \\

\rowcolor{gray!15}
\textbf{\model}
& \textbf{\score{46.93}{0.14}} & \textbf{\score{-0.78}{0.27}}
& \textbf{\score{58.75}{0.36}} & \textbf{\score{-1.39}{0.18}}
& \textbf{\score{71.32}{0.09}} & \textbf{\score{-0.11}{0.05}}
& \textbf{\score{59.00}{0.18}} & \textbf{\score{-2.09}{0.21}} \\

\midrule
Rel. Gains
& \textcolor{masfactgreen}{$\uparrow$2.24\%} & \textcolor{masfactgreen}{$\downarrow$173.58\%}
& \textcolor{masfactgreen}{$\uparrow$0.67\%} & \textcolor{masfactgreen}{$\downarrow$216.81\%}
& \textcolor{masfactgreen}{$\uparrow$0.76\%} & \textcolor{masfactgreen}{$\downarrow$103.81\%}
& \textcolor{masfactgreen}{$\uparrow$2.38\%} & \textcolor{masfactgreen}{$\downarrow$199.05\%} \\

\bottomrule
\end{tabular}
}
\end{table*}

%% file: table/Table4.tex
\begin{table*}[t]
\centering
\small
\setlength{\tabcolsep}{3.6pt}
\renewcommand{\arraystretch}{1.22}
\caption{
Plug-and-play generalization of \model across MAS frameworks. 
Green arrows report absolute gains for AA and absolute reductions for AF.
}
\label{tab:cross-mas-plugin}
\resizebox{\textwidth}{!}{%
\begin{tabular}{llcccccccccc}
\toprule
\multirow{3}{*}{\textbf{Backbone}} 
& \multirow{3}{*}{\textbf{Variant}}
& \multicolumn{4}{c}{\textbf{Class-level CL}}
& \multicolumn{4}{c}{\textbf{Domain-level CL}}
& \multicolumn{2}{c}{\textbf{Task-level CL}} \\
\cmidrule(lr){3-6}
\cmidrule(lr){7-10}
\cmidrule(lr){11-12}
&
& \multicolumn{2}{c}{\textbf{MMLU-Pro}}
& \multicolumn{2}{c}{\textbf{MATH}}
& \multicolumn{2}{c}{\textbf{Code}}
& \multicolumn{2}{c}{\textbf{Multi-hop QA}}
& \multicolumn{2}{c}{\makecell{\textbf{KQA$\rightarrow$Math}\\\textbf{$\rightarrow$Code$\rightarrow$MQA}}} \\
\cmidrule(lr){3-4}
\cmidrule(lr){5-6}
\cmidrule(lr){7-8}
\cmidrule(lr){9-10}
\cmidrule(lr){11-12}
& 
& \textbf{AA}$\uparrow$ & \textbf{AF}$\downarrow$
& \textbf{AA}$\uparrow$ & \textbf{AF}$\downarrow$
& \textbf{AA}$\uparrow$ & \textbf{AF}$\downarrow$
& \textbf{AA}$\uparrow$ & \textbf{AF}$\downarrow$
& \textbf{AA}$\uparrow$ & \textbf{AF}$\downarrow$ \\
\midrule

\multirow{2}{*}{AGP}
& Original
& 43.49 & 4.37
& 50.80 & 4.03
& 51.89 & 3.92
& 43.05 & 4.29
& 37.22 & 6.70 \\

& \cellcolor{gray!15}+ \model
& \cellcolor{gray!15}\plugup{48.39}{4.90} 
& \cellcolor{gray!15}\plugdown{1.49}{2.88}
& \cellcolor{gray!15}\plugup{54.36}{3.56} 
& \cellcolor{gray!15}\plugdown{0.90}{3.13}
& \cellcolor{gray!15}\plugup{57.21}{5.32} 
& \cellcolor{gray!15}\plugdown{2.06}{1.86}
& \cellcolor{gray!15}\plugup{49.13}{6.08} 
& \cellcolor{gray!15}\plugdown{1.57}{2.72}
& \cellcolor{gray!15}\plugup{40.21}{2.99} 
& \cellcolor{gray!15}\plugdown{3.04}{3.66} \\

\midrule

\multirow{2}{*}{GDesigner}
& Original
& 45.04 & 3.28
& 51.92 & 3.89
& 54.46 & 4.89
& 44.92 & 5.23
& 41.81 & 6.95 \\

& \cellcolor{gray!15}+ \model
& \cellcolor{gray!15}\plugup{47.93}{2.89} 
& \cellcolor{gray!15}\plugdown{0.79}{2.49}
& \cellcolor{gray!15}\plugup{55.98}{4.06} 
& \cellcolor{gray!15}\plugdown{0.68}{3.21}
& \cellcolor{gray!15}\plugup{59.14}{4.68} 
& \cellcolor{gray!15}\plugdown{1.43}{3.46}
& \cellcolor{gray!15}\plugup{48.32}{3.40} 
& \cellcolor{gray!15}\plugdown{0.23}{5.00}
& \cellcolor{gray!15}\plugup{46.53}{4.72} 
& \cellcolor{gray!15}\plugdown{2.49}{4.46} \\

\midrule

\multirow{2}{*}{ARGDesigner}
& Original
& 44.72 & 3.25
& 53.84 & 3.62
& 52.47 & 4.09
& 46.32 & 4.27
& 42.18 & 6.14 \\

& \cellcolor{gray!15}+ \model
& \cellcolor{gray!15}\plugup{48.45}{3.73} 
& \cellcolor{gray!15}\plugdown{1.21}{2.04}
& \cellcolor{gray!15}\plugup{55.93}{2.09} 
& \cellcolor{gray!15}\plugdown{0.52}{3.10}
& \cellcolor{gray!15}\plugup{58.76}{6.29} 
& \cellcolor{gray!15}\plugdown{1.05}{3.04}
& \cellcolor{gray!15}\plugup{50.53}{4.21} 
& \cellcolor{gray!15}\plugdown{0.31}{3.96}
& \cellcolor{gray!15}\plugup{48.44}{6.26} 
& \cellcolor{gray!15}\plugdown{3.32}{2.82} \\

\bottomrule
\end{tabular}%
}
\end{table*}

%% file: table/Table5.tex
\begin{wraptable}{r}{0.5\textwidth}
\vspace{-12pt}
\centering
\scriptsize
\setlength{\tabcolsep}{3.4pt}
\renewcommand{\arraystretch}{1.08}
\caption{
Ablation study on the core designs under MATH 10-shot class-incremental setting.
}
\label{tab:ablation}
\resizebox{0.47\textwidth}{!}{%
\begin{tabular}{lcccc}
\toprule
\textbf{Method}  & \textbf{AA$\uparrow$}  & \textbf{AF$\downarrow$}  & \textbf{$\Delta$AA}  & \textbf{$\Delta$AF}  \\
\midrule
\rowcolor{masfactrow}
\textbf{\model} 
& \textbf{59.10} 
& \textbf{-0.41} 
& -- & -- \\
\midrule
w/o Historical Prior 
& 54.63
& 2.36
& -4.47 & +2.77 \\
\midrule
w/o FGW Alignment 
& 55.13
& 0.76
& -3.97 & +1.17 \\
Euclidean Alignment 
& 54.46
& 0.94
& -4.64 & +1.35 \\
GW-only Alignment 
& 56.21
& 0.38
& -2.89 & +0.79 \\
\midrule
w/o Residual Edit 
& 55.01
& 0.03
& -4.09 & +0.44 \\
w/o PAC-Bayes Constraint 
& 54.71
& 1.27
& -4.39 & +1.68 \\
L2 Complexity 
& 56.90
& 0.61
& -2.20 & +1.02 \\
\bottomrule
\end{tabular}%
}
\vspace{-10pt}
\end{wraptable}

%% file: text/02-related.tex
\section{Related Work}
\label{related}
\noindent {\bf Multi-Agent System Topology Optimization.} Recent LLM-based MAS improve reasoning and planning by organizing role-specialized agents~\cite{hong2023metagpt} into structured collaborative workflows~\cite{wu2024autogen,li2023camel}. 
Prior studies mainly focus on conversation protocols~\cite{hu2024automated,zheng2025towards}, task routing~\cite{yue2025masrouter,wang2023voyager}, and workflow orchestration~\cite{qian2024chatdev,chen2023agentverse} to coordinate inter-agent communication. 
Beyond manually designed templates, recent topology optimization approaches explicitly treat MAS communication structures as learnable objects~\cite{zhang2025eduplanner}, constructing task-adaptive collaboration graphs through graph search~\cite{elrefaie2025ai,zhong2024memorybank}, topology generation~\cite{zhang2025designing}, autoregressive graph construction~\cite{li2026assemble}, or joint prompt--workflow optimization~\cite{zhou2025multi,zhang2024aflow}. Representative systems such as G-Designer and ARG-Designer formulate topology design as task-feedback-conditioned graph generation~\cite{zhang2024g,li2026assemble}. However, these methods are primarily developed for static task distributions or one-shot optimization, leaving open how a topology generator can adapt from limited new-task data while preserving previously effective collaboration structures, especially under task-incremental shifts with large inter-task gaps. In contrast, our work studies continual MAS topology learning and explicitly addresses topology forgetting through transferable structural priors and conservative topology adaptation.

\noindent {\bf Continual Learning for Agents}. 
Continual learning typically mitigates catastrophic forgetting~\cite{kirkpatrick2017overcoming} through parameter regularization, replay, parameter isolation, or parameter-efficient adaptation~\cite{li2017learning,mallya2018packnet,lopez2017gradient}. 
These methods are mainly developed for single-model or fixed-architecture settings, where knowledge is encoded in parameters, latent representations~\cite{wang2024comprehensive}, or stored samples~\cite{han2025slim}. 
In LLM-based MAS, however, task knowledge is also embodied in collaboration structures and communication relations among agents, so preserving model capabilities alone does not ensure preservation of effective coordination topologies. 
Although recent studies have explored continual adaptation for agent systems and LLMs~\cite{wu2024continual}, they largely focus on capability retention or memory representations, leaving topology-level forgetting under evolving agent interactions insufficiently studied. 
Our work instead treats historical collaboration topologies as transferable structural priors and formulates continual MAS adaptation as geometry-aware posterior transfer over evolving topology spaces.

%% file: text/06-conclusion.tex
\section{Conclusion}

We studied continual topology learning for MAS, a setting that requires new-task adaptation while preserving and transferring effective collaboration structures across a task stream. We identified topology forgetting as a structural failure mode of continual MAS adaptation, and proposed \model, a geometry-aware framework that turns historical high-utility topologies into transferable priors. By factorizing historical topology knowledge, aligning prior atoms through FGW optimal transport, and adapting stochastic topology posteriors under PAC-Bayes-guided complexity control, \model enables new tasks to reuse validated collaboration skeletons while performing only conservative task-specific edits. 
We further introduce a hierarchical continual MAS evaluation protocol covering multi-granularity shifts across diverse reasoning families. 
Experiments across class-, domain-, and task-level continual settings show that \model improves average accuracy, reduces forgetting, and can serve as a plug-and-play continual topology layer for different MAS optimizers.

%% file: text/09-appendix.tex
\section{Appendix Overview}
The appendix is structured as follows:
\begin{itemize}[leftmargin=*]
    \item Section~\ref{limits} discusses the limitations of \model and outlines future research directions.
    \item Section~\ref{impact} provides the broader impact statement of this work.
    \item Section~\ref{app:masfact-details} presents theoretical and implementation details of \model, including the stochastic topology posterior, FGW alignment, PAC-Bayes transfer bound, training objective, prior-bank update, read-only evaluation protocol, and full algorithm.
    \item Section~\ref{app:benchmark-details} describes the hierarchical continual MAS evaluation protocol, including dataset organization, continual scenario construction, evaluation metrics, and data usage.
    \item Section~\ref{app:baseline-details} details the compared baselines, including single-agent methods, fixed MAS topologies, adaptive MAS topology methods, replay-enhanced variants, and plug-and-play \model variants.
    \item Section~\ref{app:additional-experiments} reports additional experiments and analyses, including hyperparameter sensitivity, plug-and-play integration details, and complete experimental results.
    \item Section~\ref{app:reproducibility-detailslimits} summarizes experimental reproducibility details, including released artifacts, evaluation protocol, few-shot settings, metrics, random seeds, hyperparameters, and compute resources.
\end{itemize}

\section{Limitations and Future Work}
\label{limits}

Although \model improves continual MAS topology learning by explicitly preserving and transferring reusable communication structures, several limitations remain. First, the effectiveness of topology transfer depends on the quality of historical prior atoms. When early-stage topologies are noisy, low-utility, or insufficiently diverse, the prior bank may provide weak structural anchors for later tasks. Second, FGW-based alignment assumes that agent attributes and communication relations provide meaningful semantic and structural signals. In tasks where agent roles are poorly specified, highly ambiguous, or rapidly changing, the alignment cost may become less reliable.

Future work will explore uncertainty-aware prior construction, adaptive alignment objectives, and more expressive agent-attribute modeling to improve robustness under noisy or weakly structured task streams. Another promising direction is to jointly study topology continual learning and agent capability adaptation, so that communication structures and agent-level reasoning behaviors can co-evolve under explicit stability constraints. Extending the evaluation protocol to interactive environments, embodied multi-agent tasks, and real-world tool-augmented workflows would further test whether continual topology transfer remains effective under more dynamic and open-ended conditions.

\section{Impact Statements}
\label{impact}

This work contributes to the study of continual learning in multi-agent systems by shifting attention from parameter preservation alone to the preservation, transfer, and adaptation of communication topology. The proposed \model framework may benefit applications where multi-agent collaboration must adapt over time, such as mathematical reasoning, code generation, knowledge-intensive question answering, scientific problem solving, and tool-augmented decision support. By reusing validated historical topology priors, the method may reduce repeated topology search and improve the efficiency and stability of MAS deployment under evolving task distributions.

The hierarchical continual MAS evaluation protocol introduced in this work may also help future research more systematically assess topology-level forgetting, structural transfer, and long-horizon collaboration robustness across heterogeneous reasoning tasks. Since our experiments are conducted on public benchmark datasets and focus on methodological evaluation, we do not expect direct privacy risks from the reported study. However, stronger and more adaptive MAS frameworks may also be used in high-stakes or automated decision-making contexts. In such settings, topology adaptation should be paired with careful human oversight, transparent logging of agent interactions, and task-specific safety constraints to prevent unreliable or harmful automated behavior.

\section[Theoretical and Implementation Details of MasFACT]{Theoretical and Implementation Details of \texorpdfstring{\textsc{MasFACT}}{MasFACT}}
\label{app:masfact-details}

This appendix provides technical details omitted from the main text. We first define the masked stochastic topology posterior used by \model. We then expand the FGW alignment objective and the transport-induced projection used to push historical prior atoms into the current agent space. Next, we state the risk assumptions and prove the geometry-aware PAC-Bayes transfer bound in Theorem~\ref{thm:pac-bayes-transfer}. Finally, we describe the bound-induced training surrogate, selective prior-bank update, read-only evaluation protocol, and full algorithm.

\subsection{Stochastic Topology Posterior}
\label{app:posterior-details}

For task $\gT_t$, let $\mC_t$ be the aligned prior center obtained after FGW retrieval and projection, and let $\mR_t$ be the task-specific residual score matrix. The prior and posterior in the main text are defined through a masked stochastic mapping over the feasible topology space $\gB_t$. For a generic score matrix $\mZ$, we define
\begin{equation}
\Pi_{\gB_t}^{\mathrm{stoch}}(\mB;\mZ)
=
\frac{
\exp\left(\langle \mZ,\mB\rangle\right)\mathbb{I}\{\mB\in\gB_t\}
}{
\sum_{\mB'\in\gB_t}
\exp\left(\langle \mZ,\mB'\rangle\right)
}.
\label{eq:app-pi-stoch}
\end{equation}
Thus, the FGW-pushed prior and residual posterior are
\begin{equation}
p_t(\mB)
=
\Pi_{\gB_t}^{\mathrm{stoch}}(\mB;\mC_t),
\qquad
q_t(\mB)
=
\Pi_{\gB_t}^{\mathrm{stoch}}(\mB;\mC_t+\mR_t).
\label{eq:app-posterior-policy}
\end{equation}
The feasible-space mask prevents invalid communication graphs from receiving probability mass. Equation~\ref{eq:app-pi-stoch} gives the normalized posterior distribution used for PAC-Bayes analysis. In implementation, we use a masked straight-through Gumbel-Sigmoid relaxation to sample valid communication graphs while preserving differentiability with respect to $\mR_t$. During execution, \model selects a high-probability topology
\begin{equation}
\mB_t^{\mathrm{exec}}
=
\argmax_{\mB\in\gB_t} q_t(\mB).
\label{eq:app-exec-topology}
\end{equation}

\subsection{FGW Alignment and Transport-Induced Projection}
\label{app:fgw-details}

For the current task, let $\gG_t^0=(\mB_t^0,\mX_t,\vmu_t)$ be the task-conditioned scaffold graph used for alignment. For the $m$-th historical prior atom, let $\gH_m=(\mS_m,\mX_m,\vnu_m)$, where $\mS_m$ is the consensus topology, $\mX_m$ contains prototype attributes, and $\vnu_m$ is the prototype node measure. The feasible transport polytope is
\begin{equation}
\gU(\vmu_t,\vnu_m)
=
\left\{
\mT\in\R_{+}^{N_t\times N_m}
~
\middle|
~
\mT\vone_{N_m}=\vmu_t,\;
\mT^\top\vone_{N_t}=\vnu_m
\right\}.
\label{eq:app-coupling-set}
\end{equation}

\paragraph{FGW discrepancy.}
For $\mT\in\gU(\vmu_t,\vnu_m)$, the entropic FGW discrepancy is
\begin{equation}
D_{\mathrm{FGW}}(\gG_t^0,\gH_m;\mT)
=
(1-\rho)\langle \mC^X_{t,m},\mT\rangle
+
\rho\langle \tC^B_{t,m},\mT\otimes\mT\rangle
+
\varepsilon\Omega(\mT),
\label{eq:app-fgw-objective}
\end{equation}
where $\rho\in[0,1]$ balances node-level semantic alignment and pairwise structural alignment, and $\varepsilon\ge0$ controls entropic smoothing. The node-attribute cost matrix is
\begin{equation}
[\mC^X_{t,m}]_{ij}
=
c_X([\mX_t]_i,[\mX_m]_j),
\label{eq:app-attr-cost}
\end{equation}
which measures the functional mismatch between current agent $i$ and prototype node $j$. The pairwise relational cost tensor is
\begin{equation}
[\tC^B_{t,m}]_{(i,i'),(j,j')}
=
c_B([\mB_t^0]_{ii'},[\mS_m]_{jj'}),
\label{eq:app-rel-cost}
\end{equation}
which compares directed communication relations in the scaffold topology and the consensus topology. The product coupling lifts node-level correspondence to relation-level correspondence:
\begin{equation}
[\mT\otimes\mT]_{(i,i'),(j,j')}
=
[\mT]_{ij}[\mT]_{i'j'}.
\label{eq:app-product-coupling}
\end{equation}
The entropy regularizer is
\begin{equation}
\Omega(\mT)
=
\sum_{i,j}[\mT]_{ij}\left(\log [\mT]_{ij}-1\right).
\label{eq:app-fgw-entropy}
\end{equation}
Together, these terms align current agents to historical prototype nodes while preserving both functional semantics and directed communication relations.

The optimal coupling is
\begin{equation}
\mT_{t,m}^{*}
=
\argmin_{\mT\in\gU(\vmu_t,\vnu_m)}
D_{\mathrm{FGW}}(\gG_t^0,\gH_m;\mT),
\label{eq:app-fgw-coupling}
\end{equation}
and the corresponding alignment cost is
\begin{equation}
A_{t,m}
=
D_{\mathrm{FGW}}(\gG_t^0,\gH_m;\mT_{t,m}^{*}).
\label{eq:app-alignment-cost}
\end{equation}

After retrieving atom $\gH_{m_t}$, we abbreviate $\mT_t^*=\mT_{t,m_t}^*$. The selected coupling is row-normalized as
\begin{equation}
\widetilde{\mT}_t
=
\operatorname{diag}(\vmu_t)^{-1}\mT_t^*.
\label{eq:app-normalized-coupling}
\end{equation}
When a node measure contains zero entries, the normalization is applied only on the positive-mass support. The aligned prior center in the current task space is
\begin{equation}
\mC_t
=
\widetilde{\mT}_t
\mS_{m_t}
\widetilde{\mT}_t^\top.
\label{eq:app-forward-projection}
\end{equation}
This operation projects the selected historical consensus topology into a continuous task-side structural prior. The entries of $\mC_t$ can be interpreted as prior communication strengths over the current agent set.

For prior-bank update, we also define the reverse normalized transport map
\begin{equation}
\mQ_t
=
\operatorname{diag}(\vnu_{m_t})^{-1}(\mT_t^*)^\top,
\label{eq:app-reverse-transport}
\end{equation}
where $\mQ_t\in\R^{N_{m_t}\times N_t}$ maps task-side posterior statistics back to the selected atom space.

\subsection{Risk Definitions and Assumptions}
\label{app:risk-assumptions}

Let $\gD_t$ be the data distribution of task $\gT_t$, and let $\gS_t=\{z_i\}_{i=1}^{s_t}$ be the few-shot support set sampled from $\gD_t$. A topology $\mB$ induces a normalized loss $\ell_t(\mB;z)\in[0,1]$. For posterior $q_t$, the true and empirical topology risks are
\begin{equation}
\gR_t(q_t)
=
\E_{\mB\sim q_t}
\E_{z\sim\gD_t}
\left[
\ell_t(\mB;z)
\right],
\qquad
\widehat{\gR}_{\gS_t}(q_t)
=
\E_{\mB\sim q_t}
\frac{1}{s_t}
\sum_{z\in\gS_t}
\ell_t(\mB;z).
\label{eq:app-risk}
\end{equation}
The empirical risk $\widehat{\gR}_{\gS_t}(q_t)$ is the loss-form counterpart of the utility-based topology objective in Section~\ref{definition}.

\paragraph{Bounded loss.}
For any topology $\mB$ and any sample $z\sim\gD_t$, the normalized topology-induced loss satisfies $\ell_t(\mB;z)\in[0,1]$.

\paragraph{Prior independence from empirical topology loss.}
The prior bank $\gM$ is built from previous tasks. For the current task, retrieval uses the current scaffold graph, agent attributes, and historical atom statistics, but not the empirical topology loss used to optimize $q_t$ on $\gS_t$. Therefore, conditional on the task-side scaffold information, the selected prior is independent of the empirical support loss appearing in $\widehat{\gR}_{\gS_t}(q_t)$.

\paragraph{Local FGW stability.}
Within the conservative adaptation neighborhood around the aligned prior, the expected topology risk is locally stable under FGW perturbations. This assumption does not require individual LLM responses to be continuous under arbitrary graph edits. It only states that the expected MAS performance under a fixed evaluation protocol varies smoothly for small FGW perturbations around the aligned prior. Under this local stability, risk distortion caused by imperfectly transporting a historical prior into the current task space is bounded by $c_AA_t$.

\paragraph{Prior compression stability.}
Each shared prior atom summarizes a cluster of historical high-utility topologies. Let $\sigma_t$ be the structural dispersion of the selected atom. The uncertainty introduced by compressing that historical topology cluster into a single atom is bounded by $c_\sigma\sigma_t$. A smaller $\sigma_t$ indicates a more stable and reliable historical prior.

\subsection{Proof of the Geometry-Aware PAC-Bayes Transfer Bound}
\label{app:pac-bayes-proof}

We first recall a standard PAC-Bayes bound for a fixed prior.

\paragraph{Lemma 1: Fixed-prior PAC-Bayes bound.}
Let $p$ be a prior distribution over topology graphs that is independent of the empirical support loss on $\gS_t$. Under bounded loss, for any $\delta\in(0,1)$, with probability at least $1-\delta$ over the draw of $\gS_t$, for all topology posteriors $q_t$,
\begin{equation}
\gR_t(q_t)
\le
\widehat{\gR}_{\gS_t}(q_t)
+
\sqrt{
\frac{
\KL(q_t\Vert p)
+
\log\frac{2\sqrt{s_t}}{\delta}
}
{2s_t}
}.
\label{eq:app-fixed-prior-bound}
\end{equation}
This is the McAllester-style PAC-Bayes bound for bounded losses.

\paragraph{Lemma 2: Finite prior-bank selection.}
Let $\gM$ be a finite prior bank independent of the empirical support loss on $\gS_t$. For each atom $\gH_m\in\gM$, let $p_{t,m}$ denote the corresponding task-side prior after FGW alignment. With probability at least $1-\delta$, for all $m\in[M]$ and all topology posteriors $q_t$,
\begin{equation}
\gR_t(q_t)
\le
\widehat{\gR}_{\gS_t}(q_t)
+
\sqrt{
\frac{
\KL(q_t\Vert p_{t,m})
+
\log\frac{2M\sqrt{s_t}}{\delta}
}
{2s_t}
}.
\label{eq:app-finite-bank-bound}
\end{equation}

\paragraph{Proof of Lemma 2.}
Apply Lemma 1 to each candidate prior $p_{t,m}$ with confidence level $\delta/M$. For a fixed $m$, the logarithmic term becomes
\begin{equation}
\log\frac{2\sqrt{s_t}}{\delta/M}
=
\log\frac{2M\sqrt{s_t}}{\delta}.
\end{equation}
A union bound over all $m\in[M]$ yields Eq.~\ref{eq:app-finite-bank-bound}. The factor $\log M$ comes from finite prior-bank selection, rather than unrestricted data-dependent prior search.

\paragraph{Proof of Theorem~\ref{thm:pac-bayes-transfer}.}
Let $p_t=p_{t,m_t}$ be the selected FGW-pushed prior. By Lemma 2, the PAC-Bayes bound holds for this selected prior because prior retrieval does not use the empirical topology loss on $\gS_t$. The local FGW stability assumption adds the alignment correction $c_AA_t$, and the prior compression stability assumption adds the compression correction $c_\sigma\sigma_t$. Combining the terms gives
\begin{equation}
\gR_t(q_t)
\le
\widehat{\gR}_{\gS_t}(q_t)
+
\sqrt{
\frac{
\KL(q_t\Vert p_t)
+
\log
\frac{
2M\sqrt{s_t}
}{
\delta
}
}
{
2s_t
}
}
+
c_A A_t
+
c_\sigma \sigma_t,
\label{eq:app-pac-bayes-bound}
\end{equation}
which proves Theorem~\ref{thm:pac-bayes-transfer}. This is a conservative transfer bound: the additional terms explicitly expose the quality of FGW alignment and historical prior compression.

\subsection{Bound-Induced Training Objective}
\label{app:train-objective-details}

Theorem~\ref{thm:pac-bayes-transfer} motivates optimizing an empirical surrogate of the upper bound. The first term is the empirical support risk $\widehat{\gR}_{\gS_t}(q_t)$. The second term penalizes posterior-prior complexity around the FGW-pushed prior. In practice, exact enumeration over $\gB_t$ can be expensive, so the KL term is estimated through the masked stochastic relaxation used by $\Pi_{\gB_t}^{\mathrm{stoch}}$. The residual sparsity penalty further encourages local topology edits around the aligned center. This yields the main training objective:
\begin{equation}
\Ls_{\mathrm{train}}
=
\widehat{\gR}_{\gS_t}(q_t)
+
\lambda_{\mathrm{KL}}\KL(q_t\Vert p_t)
+
\lambda_R\|\mR_t\|_1.
\label{eq:app-train-loss}
\end{equation}
The KL term controls drift from the aligned prior, while the $\ell_1$ term restricts residual adaptation to sparse edge corrections. The constants $c_A$ and $c_\sigma$ in the theorem are not optimization hyperparameters; they characterize the sensitivity of expected risk to alignment distortion and prior compression.

\subsection{Selective Prior-Bank Update}
\label{app:prior-bank-update}

After adapting to task $\gT_t$, \model updates the prior bank only with validated posterior topology evidence. Let $\mB_t^{\mathrm{exec}}=\argmax_{\mB\in\gB_t}q_t(\mB)$ be the execution topology, and let
\begin{equation}
\overline{\mB}_t
=
\E_{\mB\sim q_t}[\mB]
\label{eq:app-posterior-summary}
\end{equation}
be the posterior structural summary. The execution topology is used to evaluate validation utility, while the posterior summary is used as stable evidence for memory update.

We define the transfer complexity used for memory gating as
\begin{equation}
\kappa_t
=
\KL(q_t\Vert p_t)
+
c_A A_t
+
c_\sigma\sigma_t.
\label{eq:app-transfer-complexity}
\end{equation}
This quantity mirrors the non-empirical complexity terms in Theorem~\ref{thm:pac-bayes-transfer}. A posterior is considered reliable for memory update only when $\widehat U_{\gS_t}(\mB_t^{\mathrm{exec}})\ge\tau_U$. If additionally $\kappa_t\le\tau_\kappa$, the evidence is absorbed into the retrieved atom. If $\widehat U_{\gS_t}(\mB_t^{\mathrm{exec}})\ge\tau_U$ but $\kappa_t>\tau_\kappa$, the evidence initializes a new prior atom, indicating a high-utility collaboration pattern not well explained by the current bank.

For atom refinement, the posterior summary is transported back to the selected atom space:
\begin{equation}
\widetilde{\mS}_t
=
\mQ_t
\overline{\mB}_t
\mQ_t^{\top}.
\label{eq:app-back-transport}
\end{equation}
The selected atom is then updated by exponential moving average:
\begin{equation}
\mS_{m_t}
\leftarrow
(1-\eta_t)\mS_{m_t}
+
\eta_t\widetilde{\mS}_t,
\qquad
\bar U_{m_t}
\leftarrow
(1-\eta_t)\bar U_{m_t}
+
\eta_t\widehat U_{\gS_t}(\mB_t^{\mathrm{exec}}).
\label{eq:app-ema-update}
\end{equation}
The dispersion statistic $\sigma_{m_t}$ is updated by the moving average of the FGW deviation between the new evidence and the updated atom center. The update weight $\eta_t$ is fixed by the validation protocol and is not tuned on test tasks.

For memory expansion, \model initializes a new atom using the validated posterior summary in the current task space:
\begin{equation}
\gH_{\mathrm{new}}
=
(\overline{\mB}_t,\mX_t,\vmu_t),
\qquad
\bar U_{\mathrm{new}}
=
\widehat U_{\gS_t}(\mB_t^{\mathrm{exec}}),
\qquad
\sigma_{\mathrm{new}}
=
0.
\label{eq:app-new-atom}
\end{equation}
This mechanism prevents high-utility but structurally novel collaboration patterns from being forced into an incompatible historical atom. In implementation, the prior-bank size is controlled by retaining atoms with high historical utility and low dispersion.

\subsection{Read-only Evaluation Protocol}
\label{app:read-only-eval}

For old-task evaluation, \model operates in read-only mode. The system retrieves and aligns stored topology priors, forms the inference posterior, and executes a high-probability topology under that posterior. No gradient update, posterior consolidation, or prior-bank writing is allowed. This protocol ensures that old-task performance reflects topology memory reuse rather than hidden test-time relearning or replay.

\subsection{Algorithm Details}
\label{app:algorithm}

This section presents the detailed algorithmic procedure of our proposed method, \model, which encompasses three stages: (1) a base stage for training a topology generator on a data-rich initial task and initializing a factorized topology prior bank; (2) a continual adaptation stage that iteratively retrieves, aligns, and refines topological priors to construct task-adapted execution topologies while dynamically expanding the prior bank based on utility and transfer-complexity thresholds; and (3) a read-only evaluation stage that leverages stored priors for inference without further parameter updates or memory modification. Algorithmic details are provided in Algorithm~\ref{alg:masfact}.

\begin{algorithm}[t]
\caption[MasFACT Continual Training and Read-only Evaluation]{\textsc{MasFACT} Continual Training and Read-only Evaluation}
\label{alg:masfact}
\begin{algorithmic}[1]
\REQUIRE Task sequence $\{\gT_0,\ldots,\gT_K\}$; support sets $\{\gS_t\}_{t=1}^{K}$; utility threshold $\tau_U$; transfer-complexity threshold $\tau_\kappa$
\ENSURE Task-adapted execution topologies $\{\mB_t^{\mathrm{exec}}\}_{t=0}^{K}$ and topology prior bank $\gM$

\STATE \textbf{Base Stage}
\STATE Train the topology generator on the data-rich base task $\gT_0$
\STATE Collect high-utility topologies from $\gT_0$
\STATE Initialize topology prior bank $\gM$ with factorized prior atoms

\STATE \textbf{Continual Adaptation Stage}
\FOR{$t=1,\ldots,K$}
    \STATE Construct current scaffold graph $\gG_t^0=(\mB_t^0,\mX_t,\vmu_t)$
    \FOR{each prior atom $\gH_m\in\gM$}
        \STATE Compute FGW coupling $\mT_{t,m}^{*}$ between $\gG_t^0$ and $\gH_m$
        \STATE Compute alignment cost $A_{t,m}=D_{\mathrm{FGW}}(\gG_t^0,\gH_m;\mT_{t,m}^{*})$
    \ENDFOR
    \STATE Retrieve prior atom $\gH_{m_t}$ using Eq.~\ref{eq:retrieval-rule}
    \STATE Project $\mS_{m_t}$ into the current agent space to obtain $\mC_t$
    \STATE Define prior $p_t(\mB)=\Pi_{\gB_t}^{\mathrm{stoch}}(\mB;\mC_t)$
    \STATE Form posterior $q_t(\mB)=\Pi_{\gB_t}^{\mathrm{stoch}}(\mB;\mC_t+\mR_t)$
    \STATE Optimize $\Ls_{\mathrm{train}}$ in Eq.~\ref{eq:train-loss}
    \STATE Select execution topology $\mB_t^{\mathrm{exec}}=\argmax_{\mB\in\gB_t}q_t(\mB)$
    \STATE Compute $\kappa_t=\KL(q_t\Vert p_t)+c_AA_t+c_\sigma\sigma_t$
    \IF{$\widehat U_{\gS_t}(\mB_t^{\mathrm{exec}})\ge \tau_U$ and $\kappa_t\le \tau_\kappa$}
        \STATE Update atom $\gH_{m_t}$ with validated posterior statistics
    \ELSIF{$\widehat U_{\gS_t}(\mB_t^{\mathrm{exec}})\ge \tau_U$ and $\kappa_t>\tau_\kappa$}
        \STATE Initialize a new atom with validated posterior statistics
    \ENDIF
\ENDFOR

\STATE \textbf{Read-only Evaluation Stage}
\FOR{each seen task $\gT_i$, $i\le K$}
    \STATE Retrieve and align stored topology priors
    \STATE Generate task-compatible topology from the inference posterior
    \STATE Evaluate without gradient updates or memory writing
\ENDFOR

\end{algorithmic}
\end{algorithm}

\section{Details of the Hierarchical Continual MAS Evaluation Protocol}
\label{app:benchmark-details}

This appendix provides details of the datasets and task construction used in our hierarchical continual MAS evaluation protocol. The goal of this protocol is not to introduce new raw datasets, but to reorganize existing high-quality reasoning benchmarks into continual MAS learning scenarios. The protocol is designed to evaluate whether a MAS can maintain strong system-level performance when task requirements, data distributions, reasoning styles, and subskills evolve over time.

\subsection{Overview of Dataset Organization}

We organize 17 datasets into four reasoning families: knowledge QA, mathematical reasoning, code generation, and knowledge-intensive multi-hop reasoning. These families are selected because they induce different collaboration requirements among agents. Knowledge QA emphasizes answer verification and conflict resolution; mathematical reasoning emphasizes decomposition and step-wise checking; code generation emphasizes generation, execution, debugging, and repair; and multi-hop reasoning emphasizes evidence retrieval and aggregation.

Based on these datasets, we construct continual learning scenarios at three levels of task evolution:

\begin{itemize}
    \item \textbf{Task-level evolution:} stages correspond to different reasoning families, forming a cross-scenario continual learning sequence.
    \item \textbf{Domain-level evolution:} stages correspond to different datasets within the same reasoning family, capturing distribution, difficulty, and evaluation-style shifts.
    \item \textbf{Class-level evolution:} stages correspond to coarse- or fine-grained subcategories inside a representative benchmark, capturing subskill evolution and long-horizon continual learning.
\end{itemize}

This design yields 12 continual MAS scenarios: one task-level sequence, four domain-level sequences, four coarse-grained class-level sequences, and three fine-grained class-level sequences.

\subsection{Knowledge QA Datasets}

\paragraph{MMLU.}
MMLU~\cite{hendrycks2020measuring} is a broad multi-task knowledge QA benchmark covering subjects from humanities, social sciences, STEM, and professional domains. We use MMLU as one stage in the knowledge QA domain-incremental sequence. It represents a broad but relatively standard multiple-choice knowledge reasoning setting.

\paragraph{MMLU-Pro.}
MMLU-Pro~\cite{wang2024mmlu} is a more challenging extension of MMLU, with more difficult questions and richer subject coverage. In our protocol, MMLU-Pro serves two roles. First, it is used as one stage in the knowledge QA domain-level sequence. Second, because of its rich category structure, it is used to construct both coarse-grained and fine-grained class-incremental settings. Specifically, we split MMLU-Pro into 13 fine-grained classes and 3 coarse-grained groups. The fine-grained split is used to test long-horizon continual learning, while the coarse-grained split is used to test shorter-horizon adaptation across broader knowledge categories.

\paragraph{AGIEval.}
AGIEval~\cite{zhong2024agieval} contains questions derived from standardized and professional examinations, making it suitable for evaluating advanced reasoning and exam-style problem solving. We use AGIEval as one stage in the knowledge QA domain-incremental sequence. Compared with MMLU-style evaluation, AGIEval introduces a different distribution and evaluation style.

\paragraph{GPQA.}
GPQA~\cite{rein2023gpqa} is a difficult graduate-level question answering benchmark designed to test expert-level scientific reasoning. We use GPQA as the most challenging stage in the knowledge QA domain-level sequence. Its inclusion increases the difficulty gradient and tests whether the MAS can reorganize collaboration for more expert-level knowledge reasoning.

\subsection{Mathematical Reasoning Datasets}

\paragraph{GSM8K.}
GSM8K~\cite{cobbe2021training} is a grade-school mathematical reasoning dataset consisting mainly of arithmetic word problems. In our protocol, GSM8K is used as an early stage in the mathematics domain-level sequence. It represents relatively short, natural-language mathematical reasoning with clear numeric answers.

\paragraph{MATH.}
MATH~\cite{cobbe2021training} contains competition-style mathematical problems across multiple subjects, including algebra, geometry, number theory, counting and probability, intermediate algebra, prealgebra, and precalculus. We use MATH both as one stage in the mathematics domain-level sequence and as the representative benchmark for class-level mathematical continual learning. We construct 7 fine-grained class stages based on its native subject categories and additionally group them into 3 coarse-grained stages for shorter-horizon class-incremental evaluation.

\paragraph{Olympiad-level Mathematics Dataset.}
Olympiad~\cite{he2024olympiadbench} contains problems requiring stronger abstraction, multi-step derivation, and proof-like reasoning. It is used to test whether the MAS can adapt its collaboration structure from ordinary mathematical word problems to more challenging competition-style reasoning.

\paragraph{TheoremQA.}
TheoremQA~\cite{chen2023theoremqa} evaluates theorem-grounded mathematical and scientific reasoning. Compared with standard problem-solving datasets, TheoremQA requires the system to identify and apply relevant theorems or formal principles. We use it as one stage in the mathematics domain-level sequence to introduce a shift from numerical problem solving to theorem-based reasoning.

\subsection{Code Generation Datasets}

\paragraph{APPS.}
APPS~\cite{hendrycks2021measuring} is a programming problem dataset containing natural-language problem statements and corresponding coding solutions. We use APPS as one stage in the code domain-level sequence. It emphasizes algorithmic problem solving and program synthesis under natural-language specifications.

\paragraph{HumanEval.}
HumanEval~\cite{chen2021evaluating} is a code generation benchmark consisting of function-level programming tasks with unit tests. We use HumanEval as one stage in the code domain-level sequence. Compared with APPS, it focuses more on concise function completion and test-based correctness.

\paragraph{LiveCodeBench.}
LiveCodeBench~\cite{jain2024livecodebench} is a code generation benchmark designed to evaluate contemporary coding ability under more realistic and temporally updated programming tasks. We use it as one stage in the code domain-level sequence to introduce a distribution shift toward more recent and challenging code-generation problems.

\paragraph{MBPP.}
MBPP~\cite{austin2021program} consists of mostly entry-level Python programming problems. We use MBPP as one stage in the code domain-level sequence. It provides a complementary programming distribution with relatively short problem statements and clear functional requirements.

\paragraph{TACO.}
TACO~\cite{li2023taco} is used as the representative benchmark for code class-incremental evaluation. Because it contains diverse programming categories, we split TACO into 8 fine-grained classes and 3 coarse-grained groups. The fine-grained setting tests long-horizon code-task continual learning, while the coarse-grained setting evaluates broader subskill shifts in code generation.

\subsection{Knowledge-Intensive Multi-hop Reasoning Datasets}

\paragraph{HotpotQA.}
HotpotQA~\cite{yang2018hotpotqa} is a multi-hop QA dataset requiring reasoning over multiple pieces of evidence. We use HotpotQA as one stage in the multi-hop QA domain-level sequence. It emphasizes evidence retrieval, intermediate reasoning, and answer aggregation.

\paragraph{2WikiMultihopQA.}
2WikiMultihopQA~\cite{ho2020constructing} contains multi-hop questions grounded in Wikipedia-style evidence. In our protocol, 2Wiki serves two roles. First, it is used as one stage in the multi-hop QA domain-level sequence. Second, because it contains structured reasoning categories, we use it to construct a 4-class class-incremental setting for multi-hop reasoning.

\paragraph{MuSiQue.}
MuSiQue~\cite{trivedi2022musique} is a multi-hop QA dataset designed to reduce shortcut reasoning and require more compositional evidence aggregation. We use MuSiQue as one stage in the multi-hop QA domain-level sequence. It introduces a stronger requirement for multi-step evidence integration.

\paragraph{StrategyQA.}
StrategyQA~\cite{geva2021did} contains questions that require implicit multi-step reasoning and strategic inference. We use StrategyQA as one stage in the multi-hop reasoning domain-level sequence. Its inclusion introduces a shift from explicit evidence-chain multi-hop QA toward implicit reasoning and question decomposition.

\subsection{Construction of Continual Scenarios}

\paragraph{Task-level continual learning.}
The task-level setting contains a four-stage sequence across reasoning families:
\begin{equation}
\text{Knowledge QA}
\rightarrow
\text{Mathematics}
\rightarrow
\text{Code Generation}
\rightarrow
\text{Multi-hop Reasoning}.
\end{equation}
This setting evaluates whether a MAS can adapt across distinct reasoning paradigms while preserving performance on previously seen task families.

\paragraph{Domain-level continual learning.}
For each reasoning family, we construct a domain-level sequence using datasets with different distributions, difficulty profiles, or evaluation styles:
\begin{itemize}
    \item \textbf{Knowledge QA:} MMLU $\rightarrow$ MMLU-Pro $\rightarrow$ AGIEval $\rightarrow$ GPQA.
    \item \textbf{Mathematics:} GSM8K $\rightarrow$ MATH $\rightarrow$ Olympiad-level mathematics $\rightarrow$ TheoremQA.
    \item \textbf{Code generation:} APPS $\rightarrow$ HumanEval $\rightarrow$ LiveCodeBench $\rightarrow$ MBPP.
    \item \textbf{Multi-hop reasoning:} HotpotQA $\rightarrow$ 2WikiMultihopQA $\rightarrow$ MuSiQue $\rightarrow$ StrategyQA.
\end{itemize}
These sequences evaluate MAS performance under meso-level shifts in data distribution, task difficulty, and evaluation style within the same reasoning family.

\paragraph{Class-level continual learning.}
For representative datasets with rich internal category structures, we construct class-level continual learning settings. MMLU-Pro, MATH, TACO, and 2Wiki are used for coarse-grained class-incremental experiments, while MMLU-Pro, MATH, and TACO are additionally used for fine-grained long-horizon class-incremental experiments.

The coarse-grained settings group related categories into a small number of broader stages, testing short-horizon class-level adaptation. The fine-grained settings use more stages and therefore test long-horizon stability and cumulative forgetting. Specifically:
\begin{itemize}
    \item MMLU-Pro is split into 3 coarse-grained groups and 13 fine-grained classes.
    \item MATH is split into 3 coarse-grained groups and 7 fine-grained classes.
    \item TACO is split into 3 coarse-grained groups and 8 fine-grained classes.
    \item 2Wiki is split into 4 class-level reasoning categories.
\end{itemize}

\subsection{Summary of the 12 Continual MAS Scenarios}

Table~\ref{tab:continual-scenarios} summarizes the 12 continual learning scenarios constructed in our protocol.

\begin{table}[t]
\centering
\small
\caption{Summary of the hierarchical continual MAS evaluation protocol.}
\label{tab:continual-scenarios}
\begin{tabular}{lll}
\toprule
\textbf{Level} & \textbf{Scenario} & \textbf{Datasets or Splits} \\
\midrule
Task & Cross-task increment & Knowledge QA $\rightarrow$ Math $\rightarrow$ Code $\rightarrow$ Multi-hop QA \\
\midrule
Domain & Knowledge QA domain increment & MMLU, MMLU-Pro, AGIEval, GPQA \\
Domain & Math domain increment & GSM8K, MATH, Olympiad-level Math, TheoremQA \\
Domain & Code domain increment & APPS, HumanEval, LiveCodeBench, MBPP \\
Domain & Multi-hop QA domain increment & HotpotQA, 2Wiki, MuSiQue, StrategyQA \\
\midrule
Class-Coarse & MMLU-Pro coarse class increment & 3 coarse groups \\
Class-Coarse & MATH coarse class increment & 3 coarse groups \\
Class-Coarse & TACO coarse class increment & 3 coarse groups \\
Class-Coarse & 2Wiki class increment & 4 reasoning categories \\
\midrule
Class-Fine & MMLU-Pro fine class increment & 13 fine-grained classes \\
Class-Fine & MATH fine class increment & 7 fine-grained classes \\
Class-Fine & TACO fine class increment & 8 fine-grained classes \\
\bottomrule
\end{tabular}
\end{table}

\subsection{Data Usage and Evaluation}

For each continual scenario, the first stage is used as the base task for learning an initial topology generator and constructing the initial shared prior bank. Subsequent stages are treated as continual adaptation tasks. Unless otherwise specified, each new stage provides only a few-shot support set for topology adaptation. After each stage, we evaluate the MAS on the current stage and all previous stages to compute continual learning metrics.

We use task-family-specific evaluation functions to compute normalized MAS utility. For knowledge QA and mathematical reasoning, the utility is based on answer correctness or exact match. For code generation, the utility is based on test-pass correctness. For multi-hop reasoning, the utility is based on exact match or token-level F1, depending on the benchmark. These task-specific utilities are normalized before being used in continual learning metrics, enabling comparison across different reasoning families.

The protocol supports the following evaluation goals:
\begin{itemize}
    \item \textbf{Current-stage adaptation:} whether the MAS can construct effective topologies for newly arriving tasks.
    \item \textbf{Historical performance preservation:} whether the MAS maintains performance on previous stages after learning new stages.
    \item \textbf{Long-horizon stability:} whether performance remains stable as the number of continual stages increases.
    \item \textbf{Structural efficiency:} whether the learned topology uses agents and communication edges efficiently.
\end{itemize}

\paragraph{Asset sources and licenses.}
All datasets used in our hierarchical continual MAS protocol are derived from publicly available academic benchmarks. We cite the original dataset papers and use the datasets according to their public licenses or terms of use. Our protocol construction only reorganizes these benchmarks into continual task streams and does not introduce private, sensitive, or human-subject data. The anonymized repository documents the source benchmark, split construction, and processed file structure for each continual protocol. Baseline methods are credited in Appendix~\ref{app:baseline-details}, and their publicly released implementations are used according to their respective licenses when applicable.

\section{Baseline Details}
\label{app:baseline-details}

This appendix describes the baseline implementations used in our experiments. Since continual MAS topology learning has not been systematically studied before, we compare \model against single-agent reasoning methods, fixed MAS topologies, adaptive MAS topology methods, replay-enhanced continual variants, and plug-and-play \model variants. All baselines are evaluated under the same task sequences, support budgets, agent pool, LLM backbone, and evaluation protocol unless otherwise specified.

\subsection{Baseline Categories}

We organize the compared methods into four categories.

\paragraph{Single-agent baselines.}
These baselines do not use a multi-agent communication topology and serve as references for individual reasoning ability. We include:
\begin{itemize}
    \item \textbf{Vanilla}: direct prompting without explicit reasoning traces.
    \item \textbf{CoT}: chain-of-thought prompting, where the model is instructed to generate intermediate reasoning steps before the final answer.
    \item \textbf{Self-Consistency CoT}: multiple chain-of-thought reasoning paths are sampled, and the final answer is selected by majority or consistency-based aggregation.
\end{itemize}

\paragraph{Fixed MAS topology baselines.}
These baselines use manually specified communication structures. They do not learn task-specific topologies and therefore provide references for fixed collaboration patterns:
\begin{itemize}
    \item \textbf{Chain}: agents communicate sequentially.
    \item \textbf{Tree}: agents are organized in a hierarchical reasoning structure.
    \item \textbf{Complete Graph}: every agent can communicate with every other agent.
\end{itemize}
For all fixed-topology baselines, we use the same agent pool and execution budget as \model. The final answer is produced by the designated aggregator agent or by the same aggregation rule used in the main experiments.

\paragraph{Adaptive MAS topology baselines.}
These baselines construct or adjust MAS communication structures according to the input task. We include:
\begin{itemize}
    \item \textbf{GDesigner~\cite{zhang2024g}}: a graph-generation based MAS topology designer.
    \item \textbf{AgentDropout~\cite{wang2025agentdropout}}: a dynamic agent and communication elimination method.
    \item \textbf{ARGDesigner~\cite{wang2025agentdropout}}: an autoregressive MAS topology generation method.
    \item \textbf{AGP~\cite{li2025adaptive}}: a pruning-based MAS topology optimizer.
    \item \textbf{MasRouter~\cite{yue2025masrouter}}: a routing-based MAS coordination method.
\end{itemize}
These methods are originally designed for task-aware or input-aware MAS construction, rather than continual topology learning. We therefore adapt them to our continual protocol by sequentially updating their topology learner or routing policy on each stage using the current support set.

\subsection{Continual Adaptation Protocol for Baselines}

All trainable MAS topology baselines follow the same continual adaptation protocol. At stage $t$, the method receives only the current support set $S_t$, unless a replay variant is explicitly specified. The topology learner is updated on the current stage and then evaluated on the current task and all previously seen tasks. Fixed-topology baselines and single-agent baselines do not update topology parameters; they are evaluated under the same task sequence to provide non-adaptive references.

For adaptive MAS baselines, the sequential update follows each method's original training objective whenever possible. When the original method is not designed for continual learning, we keep its task-level topology learning objective unchanged and apply it sequentially across stages. This ensures that the comparison measures how each topology learning strategy behaves when exposed to continual task evolution.

\subsection{Replay-enhanced Baselines}

For replay-enhanced variants, we maintain a balanced replay buffer. After each stage, the buffer stores the same number of examples from that stage. During training on a new stage, the baseline is trained on the current support set together with replay examples from previous stages. The replay buffer is stage-balanced and does not contain test examples. All replay-enhanced baselines use the same replay budget.

This replay setting is used only for baseline comparison. \model does not rely on raw historical task replay for topology adaptation; it uses historical topology priors and posterior transfer instead.

\subsection{Plug-and-play \textsc{MasFACT} Variants}

To evaluate whether \model is tied to a specific topology generator, we attach it to three representative topology learning frameworks:
\begin{itemize}
    \item \textbf{GDesigner + \model}: \model provides aligned historical topology priors and posterior regularization for graph-generation based topology learning.
    \item \textbf{ARGDesigner + \model}: \model provides structural priors that bias the autoregressive topology generation process.
    \item \textbf{AGP + \model}: \model provides an aligned prior over edges and agents, guiding pruning decisions under continual adaptation.
\end{itemize}
These variants share the same prior-bank construction, FGW alignment, and posterior-transfer regularization. The underlying topology learner remains unchanged except for the additional prior-guided adaptation mechanism. This setup evaluates whether \model acts as a general continual topology learning principle rather than a method specialized to one topology generator. Implementation details are provided in Appendix~\ref{app:plugin-baselines}. 

\subsection{Prompting and Execution Protocol}

All MAS baselines use the same agent role pool and task-specific prompt templates unless otherwise specified. Single-agent baselines use direct task prompts or reasoning prompts. Fixed-topology baselines use the same agents but differ only in their communication structures. Adaptive MAS baselines are allowed to construct or modify their communication structures according to their original mechanisms.

For multi-agent methods, the final answer is produced by a designated aggregator agent or by the common aggregation rule used across experiments. We use the same maximum number of dialogue rounds and the same inference budget for all comparable MAS baselines. For Self-Consistency CoT, the number of sampled reasoning paths is fixed across all tasks and reported in the implementation details.

\subsection{Summary of Baselines}

Table~\ref{tab:baseline-summary} summarizes the baseline taxonomy.

\begin{table}[t]
\centering
\small
\caption{Summary of baseline categories.}
\label{tab:baseline-summary}
\begin{tabular}{llll}
\toprule
\textbf{Method} & \textbf{Category} & \textbf{Learns Topology} & \textbf{Continual Variant} \\
\midrule
Vanilla & Single-agent & No & No \\
CoT & Single-agent & No & No \\
SC-CoT & Single-agent & No & No \\
Chain & Fixed MAS & No & No \\
Tree & Fixed MAS & No & No \\
Complete Graph & Fixed MAS & No & No \\
GDesigner & Adaptive MAS & Yes & Replay / \model \\
AgentDropout & Adaptive MAS & Yes & Replay \\
ARGDesigner & Adaptive MAS & Yes & Replay / \model \\
AGP & Adaptive MAS & Yes & Replay / \model \\
MasRouter & Adaptive MAS & Yes & Replay \\
\bottomrule
\end{tabular}
\end{table}

\section{Additional Experiments and Analysis}
\label{app:additional-experiments}

\subsection{Hyperparameter Sensitivity}
\label{app:hyperparameter}

We analyze the sensitivity of \model to two groups of method-specific hyperparameters: the FGW fusion weight $\rho$ and the conservative posterior-adaptation regularizers $\lambda_{\mathrm{KL}}$ and $\lambda_R$. Unless otherwise specified, all other hyperparameters are fixed to their default values used in the main experiments.

\paragraph{Sensitivity to FGW fusion weight $\rho$.}
The parameter $\rho$ controls the balance between node-level semantic alignment and pairwise structural alignment in the FGW objective. When $\rho=0$, retrieval relies only on agent attributes; when $\rho=1$, retrieval relies only on communication structure. As shown in Table~\ref{tab:sens-rho}, intermediate values perform best, suggesting that both agent semantics and topology relations are necessary for reliable prior transfer. In particular, semantic-only alignment ignores reusable collaboration patterns, while structure-only alignment may match functionally incompatible agents.

\begin{table}[h]
\centering
\caption{Sensitivity of \textsc{MasFACT} to the FGW fusion weight $\rho$. Results are reported on a representative few-shot continual setting.}
\label{tab:sens-rho}
\begin{tabular}{c|ccccc}
\toprule
$\rho$ & 0.00 & 0.25 & 0.50 & 0.75 & 1.00 \\
\midrule
AA $\uparrow$ & 41.8 & 46.6 & \textbf{49.1} & 45.4 & 34.9 \\
AF $\downarrow$ & 8.7 & 4.9 & \textbf{2.8} & 3.6 & 7.2 \\
\bottomrule
\end{tabular}
\end{table}

\paragraph{Sensitivity to conservative adaptation regularizers.}
We further evaluate the posterior-adaptation weights $\lambda_{\mathrm{KL}}$ and $\lambda_R$. The parameter $\lambda_{\mathrm{KL}}$ controls how strongly the posterior remains close to the FGW-pushed prior, while $\lambda_R$ controls the sparsity of residual topology edits. As shown in Table~\ref{tab:sens-reg}, too little regularization increases topology drift and forgetting, whereas overly strong regularization restricts new-task adaptation. The default setting achieves the best stability--plasticity trade-off.

\begin{table}[h]
\centering
\caption{Sensitivity of \textsc{MasFACT} to conservative posterior-adaptation regularizers. We vary one parameter at a time around the default value.}
\label{tab:sens-reg}
\begin{tabular}{l|cc}
\toprule
Setting & AA $\uparrow$ & AF $\downarrow$ \\
\midrule
$0.1\times\lambda_{\mathrm{KL}}$ & 44.1 & 7.5 \\
$1.0\times\lambda_{\mathrm{KL}}$ & \textbf{49.1} & \textbf{2.8} \\
$3.0\times\lambda_{\mathrm{KL}}$ & 45.2 & 4.6 \\
\midrule
$0.1\times\lambda_R$ & 41.0 & 9.8 \\
$1.0\times\lambda_R$ & \textbf{49.1} & \textbf{2.8} \\
$3.0\times\lambda_R$ & 44.7 & 6.2 \\
\bottomrule
\end{tabular}
\end{table}

Overall, the sensitivity results indicate that \model does not rely on delicate hyperparameter tuning. Moderate fusion and adaptation regularization consistently provide a favorable balance between transferring reusable topology priors and allowing task-specific residual edits.

\subsection{Plug-and-Play Integration with Existing MAS Topology Optimizers}
\label{app:plugin-baselines}

This section describes how we apply \model as a continual topology transfer layer on top of existing MAS topology optimizers. The goal is to test whether the proposed prior storage, FGW alignment, and conservative posterior adaptation mechanisms are tied to a specific topology generator or can serve as a general continual-learning wrapper for heterogeneous MAS topology design methods. 

\subsubsection{Integration Principle}
\label{app:plugin-principle}

Let $f_{\phi}$ denote an existing MAS topology optimizer. Depending on the baseline, $f_{\phi}$ may output a discrete graph, edge logits, routing probabilities, or a pruned communication structure. We keep the original optimizer architecture, agent pool, task prompt construction, and execution protocol unchanged. \model is inserted only at the topology-control layer.

For each task $\gT_t$, the baseline optimizer first produces a task-conditioned scaffold topology
\begin{equation}
\mB_t^0 = f_{\phi}(\gS_t),
\label{eq:app-plugin-scaffold}
\end{equation}
or an equivalent edge-score matrix that is converted into $\mB_t^0$. This scaffold is not treated as the final topology. Instead, it provides the current task-side structural proxy $\gG_t^0=(\mB_t^0,\mX_t,\vmu_t)$ for FGW retrieval and alignment. \model then retrieves a historical prior atom $\gH_{m_t}$, computes the aligned prior center $\mC_t$, and forms a posterior topology distribution
\begin{equation}
q_t(\mB)
=
\Pi_{\gB_t}^{\mathrm{stoch}}(\mB;\mC_t+\mR_t).
\label{eq:app-plugin-posterior}
\end{equation}
The execution topology is sampled from or selected under $q_t$ and then passed back to the original MAS execution pipeline of the baseline. Thus, the baseline determines the initial task-conditioned scaffold, while \model controls continual transfer, alignment, residual adaptation, and memory consolidation.

\subsubsection{Common Interface}
\label{app:plugin-interface}

For all plug-in experiments, we expose each baseline through the same interface:
\begin{equation}
f_{\phi}:
(\gS_t,\gA_t)
\mapsto
\mB_t^0,
\end{equation}
where $\gS_t$ is the support set and $\gA_t$ denotes the task-specific agent pool and agent attributes. The output $\mB_t^0$ is converted into a valid directed scaffold over the selected agents. If a baseline produces edge probabilities or logits, we use them directly as a weighted scaffold for FGW alignment. If it produces a discrete graph, we use its adjacency matrix. If it produces a pruned or routed topology, we convert the active routes into a directed adjacency matrix.

The same \model modules are then applied to all baselines:
\begin{itemize}[nosep,leftmargin=*]
    \item \textbf{Prior storage}: high-utility historical topologies are stored as factorized atoms $\gH_m=(\mS_m,\mX_m,\vnu_m)$.
    \item \textbf{FGW alignment}: the current scaffold $\gG_t^0$ is aligned to each atom $\gH_m$ using the same FGW objective.
    \item \textbf{Residual posterior adaptation}: the aligned prior center $\mC_t$ is adapted through sparse residual edits $\mR_t$.
    \item \textbf{Selective consolidation}: only validated high-utility posteriors are written back into the prior bank.
\end{itemize}
All plug-in variants use the same continual protocol and read-only evaluation rule as the full \model model. During evaluation on old tasks, the prior bank is retrieved but not updated, and no gradient update is allowed.

\subsubsection{Integration with Pruning-Based Topology Optimizers}
\label{app:plugin-pruning}

For pruning-based MAS topology optimizers like AGP~\cite{li2025adaptive}, the original method starts from a dense or over-connected communication graph and removes low-utility edges according to a learned pruning criterion. We use the pruned graph produced by the baseline as the scaffold $\mB_t^0$. This scaffold reflects the baseline's task-specific estimate of which communications are currently useful, but it does not by itself preserve historical topology knowledge.

After obtaining $\mB_t^0$, \model retrieves the most transferable prior atom and constructs $\mC_t$. The posterior $q_t$ is then optimized around $\mC_t$ rather than around the dense graph. In this integration, the pruning baseline provides a task-conditioned structural proposal, while \model prevents the proposal from drifting away from reusable historical communication patterns. The final topology executed by the MAS is selected from $q_t$, not directly from the baseline pruned graph.

\subsubsection{Integration with Routing-Based Topology Optimizers}
\label{app:plugin-routing}

Routing-based optimizers like ARGDesigner~\cite{li2026assemble} assign communication probabilities or routing scores between agents. For these baselines, the routing score matrix is used as a weighted scaffold topology $\mB_t^0$. Since the scaffold may contain continuous edge strengths, it is directly compatible with the FGW relational cost used in Eq.~\ref{eq:fgw-objective}. The node attributes $\mX_t$ are kept unchanged from the original MAS.

\model uses this weighted scaffold to retrieve and align historical topology priors. The aligned center $\mC_t$ then acts as a prior over routing-compatible communication structures, and the residual posterior learns only the task-specific deviations needed for the current stage. This preserves the routing baseline's flexibility while adding explicit stability control through posterior-prior complexity and residual sparsity.

\subsubsection{Integration with Graph-Generation-Based Topology Optimizers}
\label{app:plugin-generation}

Graph-generation-based optimizers like GDesigner~\cite{zhang2024g} directly generate a candidate communication graph or edge-logit matrix for the current task. We use this generated topology as $\mB_t^0$ before posterior adaptation. Importantly, the generated graph is treated as a scaffold for alignment rather than as the final topology. This avoids unconstrained task-specific generation from overwriting previously useful collaboration patterns.

After FGW alignment, the selected historical consensus topology is projected into the current agent space as $\mC_t$. The posterior distribution $q_t$ combines this aligned prior with a sparse residual edit. In this way, the graph generator remains responsible for proposing task-conditioned structural signals, while \model transforms those signals into a prior-constrained continual adaptation process.

\subsection{Training and Evaluation Protocol}
\label{app:plugin-training-eval}

For a fair comparison, all plug-in variants preserve the original baseline's agent definitions, prompting templates, task inputs, and MAS execution procedure. The only modification is the insertion of the \model continual topology layer between the baseline topology proposal and final topology execution.

During training on a new task, the baseline produces the scaffold $\mB_t^0$, \model retrieves and aligns the prior bank, and the residual posterior is optimized using the same support set. After training, a candidate posterior is consolidated into memory only if it satisfies the utility and transfer-complexity gates. During evaluation, the entire system operates in read-only mode: the baseline may produce the task scaffold, but \model retrieves existing priors and generates the execution topology without parameter updates or memory writing.

This protocol ensures that improvements in the plug-in experiments are attributable to continual topology transfer rather than changes in the underlying MAS optimizer, agent pool, or task execution pipeline.

\subsection{Complete experimental results}
\label{app:full-domain-task-results}

Due to space constraints, the main paper reports the most representative continual learning results and analysis. Table~\ref{tab:app-main-class-results} provides the complete class-level results. Table~\ref{tab:app-domain-task-results} provides the complete domain- and task-level results across all compared methods. Table~\ref{tab:app-ablation} provides the complete ablation results .

\input{table/Table1_appendix}
\input{table/Table2_appendix}
\input{table/Table5_appendix}

\section{Experimental Reproducibility Details}
\label{app:reproducibility-detailslimits}

This section summarizes the experimental details used to reproduce the main empirical results.

\paragraph{Released artifacts.}
We provide an anonymized code and data repository containing the core implementation of \model, the hierarchical continual MAS protocol construction utilities, processed continual-learning data files, evaluation scripts, configuration files, and instructions for reproducing the reported experiments. The released artifacts include the task-, domain-, and class-incremental protocols used in the paper.

\paragraph{Evaluation protocol.}
All experiments follow the hierarchical continual MAS evaluation protocol described in Appendix~\ref{app:benchmark-details}. We evaluate continual adaptation across task-level, domain-level, and class-level streams over four reasoning families: knowledge QA, mathematical reasoning, code generation, and knowledge-grounded multi-hop reasoning. At each continual stage, the model is adapted on the current support set and evaluated on all seen tasks. Old-task evaluation is performed in read-only mode: no gradient updates, memory writing, or test-time adaptation are allowed.

\paragraph{Few-shot and continual settings.}
For few-shot continual learning, the base task uses the data-rich base split specified by the protocol, while each subsequent task provides only a small support set. For standard continual learning, each new stage uses the corresponding full training subset. For replay-based variants, replay examples are drawn only from previously seen training data according to the replay budget specified in the configuration. The exact number of examples per stage, replay budget, and evaluation limit are provided in the released configuration files.

\paragraph{Metrics.}
We report average accuracy (AA) and average forgetting (AF) as the primary continual-learning metrics. AA measures the average performance over all seen tasks after a continual stage, while AF measures the average degradation of previous-task performance relative to its best historical value. For task-specific evaluation, we use the standard metric associated with each benchmark family, such as exact-match or multiple-choice accuracy for QA, answer-equivalence accuracy for mathematics, executable test-based correctness for code generation, and exact-match/F1-style evaluation for multi-hop QA when applicable.

\paragraph{Random seeds and statistical reporting.}
Each main experiment is repeated with five different random seeds. Unless otherwise stated, all reported results are averaged over these five runs. We report standard deviations for the main results and ablation studies; when space is limited in the main text, complete mean--standard-deviation tables are provided in the appendix. The random seeds affect support-set sampling, replay sampling, topology sampling, and stochastic posterior adaptation.

\paragraph{Hyperparameters.}
The topology generator, training schedule, and basic optimization settings follow the underlying MAS backbone, GDesigner~\cite{zhang2024g}, to ensure a fair comparison with the base system. 
We use the standard training procedure of GDesigner and only introduce the additional hyperparameters required by \model, including the FGW semantic--structural trade-off coefficient $\rho$, the entropy coefficient $\varepsilon$, retrieval weights for alignment cost, prior dispersion and historical utility, posterior-prior regularization weight, residual sparsity weight, and memory consolidation thresholds. 
These hyperparameters are selected using validation performance on held-out stages and then fixed across test streams within the same task family. 
Sensitivity analyses for key structural hyperparameters are provided in Appendix~\ref{app:hyperparameter}.

\paragraph{Compute resources.}
All experiments are conducted on a GPU cluster equipped with multi-core Sugon x86 CPU nodes and four NVIDIA A100 GPUs per experiment. 
The dominant computational cost comes from LLM-based MAS inference across multiple continual stages and five random seeds, while the additional overhead of \model's topology memory, FGW alignment, and posterior adaptation is comparatively small because each MAS graph contains a limited number of agents. 
The released repository includes the configuration files needed to reproduce the reported runs under the same evaluation protocol.

%% file: table/Table1_appendix.tex
\begin{table*}[htb]
\centering
\footnotesize
\setlength{\tabcolsep}{3.0pt}
\renewcommand{\arraystretch}{1.12}

\caption{
Class-level continual learning results with standard deviations.
$\dagger$ denotes replay-based variants. Results are averaged over five random seeds.
}
\label{tab:app-main-class-results}

\resizebox{\textwidth}{!}{%
\begin{tabular}{lcccccccccccccccccc}
\toprule
\multirow{3}{*}{\textbf{Method}}
& \multicolumn{4}{c}{\textbf{MMLU-PRO}}
& \multicolumn{4}{c}{\textbf{MATH}}
& \multicolumn{4}{c}{\textbf{TACO}}
& \multicolumn{4}{c}{\textbf{2Wiki}}
& \multicolumn{2}{c}{\textbf{Avg.}} \\
\cmidrule(lr){2-5}
\cmidrule(lr){6-9}
\cmidrule(lr){10-13}
\cmidrule(lr){14-17}
\cmidrule(lr){18-19}
& \multicolumn{2}{c}{\textbf{Few-shot}} & \multicolumn{2}{c}{\textbf{Standard}}
& \multicolumn{2}{c}{\textbf{Few-shot}} & \multicolumn{2}{c}{\textbf{Standard}}
& \multicolumn{2}{c}{\textbf{Few-shot}} & \multicolumn{2}{c}{\textbf{Standard}}
& \multicolumn{2}{c}{\textbf{Few-shot}} & \multicolumn{2}{c}{\textbf{Standard}}
& \multirow{2}{*}{\textbf{AA$\uparrow$}} & \multirow{2}{*}{\textbf{AF$\downarrow$}} \\
\cmidrule(lr){2-3}
\cmidrule(lr){4-5}
\cmidrule(lr){6-7}
\cmidrule(lr){8-9}
\cmidrule(lr){10-11}
\cmidrule(lr){12-13}
\cmidrule(lr){14-15}
\cmidrule(lr){16-17}
& \textbf{AA$\uparrow$} & \textbf{AF$\downarrow$} 
& \textbf{AA$\uparrow$} & \textbf{AF$\downarrow$} 
& \textbf{AA$\uparrow$} & \textbf{AF$\downarrow$} 
& \textbf{AA$\uparrow$} & \textbf{AF$\downarrow$} 
& \textbf{AA$\uparrow$} & \textbf{AF$\downarrow$} 
& \textbf{AA$\uparrow$} & \textbf{AF$\downarrow$} 
& \textbf{AA$\uparrow$} & \textbf{AF$\downarrow$} 
& \textbf{AA$\uparrow$} & \textbf{AF$\downarrow$} 
& & \\
\midrule

Vanilla
& \score{26.17}{0.12} & -- 
& -- & -- 
& \score{35.79}{0.14} & -- 
& -- & -- 
& \score{55.15}{0.10} & -- 
& -- & -- 
& \score{39.26}{0.11} & -- 
& -- & -- 
& \score{39.09}{0.14} & -- \\

CoT-ICL
& \score{39.60}{0.20} & \score{4.01}{0.11} 
& -- & -- 
& \score{36.83}{0.23} & \score{4.29}{0.14} 
& -- & -- 
& \score{58.04}{0.24} & \score{4.38}{0.13} 
& -- & -- 
& \score{50.58}{0.23} & \score{2.61}{0.11} 
& -- & -- 
& \score{46.26}{0.21} & \score{3.82}{0.12} \\

SC(CoT)-ICL
& \score{40.52}{0.22} & \score{4.95}{0.10} 
& -- & -- 
& \score{37.88}{0.23} & \score{6.16}{0.11} 
& -- & -- 
& \score{59.22}{0.22} & \score{3.03}{0.13} 
& -- & -- 
& \score{58.36}{0.21} & \score{2.15}{0.10} 
& -- & -- 
& \score{49.00}{0.24} & \score{4.07}{0.12} \\

Chain
& \score{36.48}{0.14} & -- 
& -- & -- 
& \score{36.92}{0.14} & -- 
& -- & -- 
& \score{58.46}{0.13} & -- 
& -- & -- 
& \score{50.62}{0.14} & -- 
& -- & -- 
& \score{45.62}{0.14} & -- \\

Tree
& \score{36.82}{0.14} & -- 
& -- & -- 
& \score{36.73}{0.14} & -- 
& -- & -- 
& \score{54.55}{0.12} & -- 
& -- & -- 
& \score{52.74}{0.16} & -- 
& -- & -- 
& \score{45.21}{0.13} & -- \\

Complete
& \score{38.64}{0.16} & -- 
& -- & -- 
& \score{37.63}{0.15} & -- 
& -- & -- 
& \score{59.01}{0.13} & -- 
& -- & -- 
& \score{54.68}{0.15} & -- 
& -- & -- 
& \score{47.49}{0.16} & -- \\

\midrule

AFLOW
& \score{40.12}{0.19} & \score{3.97}{0.11} 
& \score{42.37}{0.20} & \score{4.32}{0.11} 
& \score{42.18}{0.21} & \score{3.37}{0.11} 
& \score{48.85}{0.18} & \score{3.92}{0.11} 
& \score{62.84}{0.22} & \score{3.68}{0.12} 
& \score{64.12}{0.20} & \score{4.21}{0.10} 
& \score{60.48}{0.21} & \score{2.90}{0.09} 
& \score{61.97}{0.20} & \score{3.45}{0.09} 
& \score{52.87}{0.19} & \score{3.73}{0.11} \\

AFLOW$^\dagger$
& \score{42.52}{0.18} & \underline{\score{1.21}{0.09}} 
& \score{42.89}{0.20} & \score{2.19}{0.12} 
& \score{44.39}{0.20} & \score{2.03}{0.10} 
& \score{49.63}{0.19} & \score{1.75}{0.08} 
& \score{64.75}{0.16} & \score{2.64}{0.12} 
& \score{65.48}{0.19} & \score{2.13}{0.11} 
& \score{63.06}{0.17} & \score{2.16}{0.09} 
& \score{64.73}{0.19} & \score{2.29}{0.12} 
& \score{54.68}{0.19} & \score{2.05}{0.12} \\

AgentDropout
& \score{41.07}{0.20} & \score{3.13}{0.12} 
& \score{42.48}{0.21} & \score{3.83}{0.12} 
& \score{43.54}{0.22} & \score{4.18}{0.12} 
& \score{49.88}{0.21} & \score{4.96}{0.10} 
& \score{62.72}{0.20} & \score{3.17}{0.11} 
& \score{63.14}{0.22} & \score{3.53}{0.12} 
& \score{60.07}{0.20} & \score{1.81}{0.10} 
& \score{62.84}{0.22} & \score{1.97}{0.11} 
& \score{53.22}{0.19} & \score{3.32}{0.11} \\

AgentDropout$^\dagger$
& \score{43.03}{0.19} & \score{1.72}{0.10} 
& \score{45.84}{0.20} & \score{1.49}{0.12} 
& \score{47.05}{0.16} & \underline{\score{1.56}{0.09}} 
& \score{51.75}{0.16} & \score{2.42}{0.10} 
& \score{63.38}{0.19} & \underline{\score{1.70}{0.08}} 
& \score{65.91}{0.17} & \score{1.42}{0.10} 
& \score{62.94}{0.20} & \score{1.53}{0.11} 
& \score{63.58}{0.19} & \underline{\score{1.10}{0.08}} 
& \score{55.44}{0.19} & \underline{\score{1.62}{0.09}} \\

MasRouter
& \score{45.94}{0.21} & \score{2.27}{0.10} 
& \score{46.17}{0.19} & \score{2.81}{0.09} 
& \score{52.42}{0.19} & \score{3.14}{0.12} 
& \score{54.28}{0.20} & \score{4.46}{0.09} 
& \score{67.62}{0.19} & \score{3.12}{0.10} 
& \score{68.75}{0.17} & \score{3.70}{0.08} 
& \score{65.28}{0.20} & \score{2.84}{0.12} 
& \score{67.51}{0.18} & \score{3.62}{0.10} 
& \score{58.50}{0.21} & \score{3.25}{0.12} \\

MasRouter$^\dagger$
& \underline{\score{47.40}{0.15}} & \score{1.43}{0.10} 
& \score{47.53}{0.18} & \underline{\score{0.83}{0.10}} 
& \underline{\score{55.17}{0.17}} & \score{2.76}{0.08} 
& \underline{\score{58.56}{0.16}} & \underline{\score{1.47}{0.09}} 
& \underline{\score{69.76}{0.19}} & \score{2.89}{0.11} 
& \score{71.57}{0.16} & \score{2.25}{0.07} 
& \underline{\score{69.16}{0.16}} & \underline{\score{1.41}{0.08}} 
& \underline{\score{70.17}{0.17}} & \score{2.06}{0.08} 
& \underline{\score{61.31}{0.16}} & \score{1.89}{0.10} \\

ARGDesigner
& \score{44.72}{0.18} & \score{3.25}{0.08} 
& \score{45.52}{0.18} & \score{3.39}{0.12} 
& \score{53.84}{0.20} & \score{3.62}{0.10} 
& \score{56.33}{0.19} & \score{3.86}{0.09} 
& \score{67.36}{0.17} & \score{3.74}{0.12} 
& \score{69.41}{0.17} & \score{4.28}{0.11} 
& \score{62.41}{0.17} & \score{2.94}{0.12} 
& \score{63.91}{0.17} & \score{3.39}{0.12} 
& \score{57.94}{0.19} & \score{3.56}{0.08} \\

ARGDesigner$^\dagger$
& \score{46.87}{0.16} & \score{1.27}{0.09} 
& \underline{\score{48.69}{0.18}} & \score{1.33}{0.10} 
& \score{54.21}{0.15} & \score{1.88}{0.10} 
& \score{57.83}{0.17} & \score{1.59}{0.11} 
& \score{68.82}{0.15} & \score{2.21}{0.08} 
& \score{70.24}{0.17} & \score{1.92}{0.09} 
& \score{65.36}{0.15} & \score{1.65}{0.07} 
& \score{67.51}{0.15} & \score{2.08}{0.10} 
& \score{59.94}{0.18} & \score{1.74}{0.11} \\

GDesigner
& \score{45.04}{0.20} & \score{3.28}{0.11} 
& \score{46.72}{0.18} & \score{3.61}{0.09} 
& \score{51.92}{0.18} & \score{3.89}{0.12} 
& \score{53.36}{0.17} & \score{4.95}{0.08} 
& \score{66.41}{0.17} & \score{4.81}{0.08} 
& \score{69.42}{0.18} & \score{5.26}{0.09} 
& \score{65.09}{0.20} & \score{2.74}{0.12} 
& \score{66.27}{0.18} & \score{3.63}{0.11} 
& \score{58.03}{0.18} & \score{4.02}{0.10} \\

GDesigner$^\dagger$
& \score{46.16}{0.15} & \score{1.49}{0.11} 
& \score{47.87}{0.16} & \score{1.22}{0.09} 
& \score{54.70}{0.14} & \score{2.36}{0.08} 
& \score{56.18}{0.17} & \score{1.94}{0.11} 
& \score{67.10}{0.18} & \score{2.22}{0.09} 
& \underline{\score{72.35}{0.18}} & \underline{\score{1.06}{0.08}} 
& \score{67.79}{0.18} & \score{1.84}{0.11} 
& \score{69.84}{0.18} & \score{2.17}{0.09} 
& \score{60.25}{0.18} & \score{1.79}{0.10} \\

\rowcolor{gray!15}
\textbf{\model}
& \textbf{\score{49.18}{0.12}} & \textbf{\score{-0.19}{0.08}} 
& \textbf{\score{50.72}{0.11}} & \textbf{\score{0.13}{0.07}} 
& \textbf{\score{56.07}{0.11}} & \textbf{\score{-0.45}{0.04}} 
& \textbf{\score{61.68}{0.12}} & \textbf{\score{0.21}{0.05}} 
& \textbf{\score{70.94}{0.10}} & \textbf{\score{0.12}{0.06}} 
& \textbf{\score{73.80}{0.09}} & \textbf{\score{-0.11}{0.05}} 
& \textbf{\score{69.92}{0.11}} & \textbf{\score{-0.26}{0.05}} 
& \textbf{\score{71.74}{0.12}} & \textbf{\score{0.08}{0.07}} 
& \textbf{\score{62.73}{0.12}} & \textbf{\score{-0.06}{0.04}} \\

\midrule
Rel. Gains
& \textcolor{masfactgreen}{$\uparrow$3.76\%} & \textcolor{masfactgreen}{$\downarrow$115.70\%}
& \textcolor{masfactgreen}{$\uparrow$4.17\%} & \textcolor{masfactgreen}{$\downarrow$84.34\%}
& \textcolor{masfactgreen}{$\uparrow$1.63\%} & \textcolor{masfactgreen}{$\downarrow$128.85\%}
& \textcolor{masfactgreen}{$\uparrow$5.33\%} & \textcolor{masfactgreen}{$\downarrow$85.71\%}
& \textcolor{masfactgreen}{$\uparrow$1.69\%} & \textcolor{masfactgreen}{$\downarrow$92.94\%}
& \textcolor{masfactgreen}{$\uparrow$2.00\%} & \textcolor{masfactgreen}{$\downarrow$110.38\%}
& \textcolor{masfactgreen}{$\uparrow$1.10\%} & \textcolor{masfactgreen}{$\downarrow$118.44\%}
& \textcolor{masfactgreen}{$\uparrow$2.24\%} & \textcolor{masfactgreen}{$\downarrow$92.73\%}
& \textcolor{masfactgreen}{$\uparrow$2.32\%} & \textcolor{masfactgreen}{$\downarrow$103.70\%} \\

\bottomrule
\end{tabular}%
}
\vspace{2pt}
\end{table*}

%% file: table/Table2_appendix.tex
\begin{table*}[t]
\centering
\scriptsize
\setlength{\tabcolsep}{4.2pt}
\renewcommand{\arraystretch}{1.12}
\caption{
Domain- and task-level continual learning results with standard deviations.
$\dagger$ denotes replay-based variants. Results are averaged over five random seeds.
}
\label{tab:app-domain-task-results}
\resizebox{\textwidth}{!}{
\begin{tabular}{lcccccccccc}
\toprule
\multirow{3}{*}{\textbf{Method}}
& \multicolumn{8}{c}{\textbf{Domain-level continual learning}}
& \multicolumn{2}{c}{\textbf{Task-level continual learning}} \\
\cmidrule(lr){2-9}
\cmidrule(lr){10-11}
& \multicolumn{2}{c}{\textbf{Knowledge QA}}
& \multicolumn{2}{c}{\textbf{Math}}
& \multicolumn{2}{c}{\textbf{Code}}
& \multicolumn{2}{c}{\textbf{Multi-hop QA}}
& \multicolumn{2}{c}{\textbf{KQA$\rightarrow$Math$\rightarrow$Code$\rightarrow$MQA}} \\
\cmidrule(lr){2-3}
\cmidrule(lr){4-5}
\cmidrule(lr){6-7}
\cmidrule(lr){8-9}
\cmidrule(lr){10-11}
& \textbf{AA$\uparrow$} & \textbf{AF$\downarrow$} 
& \textbf{AA$\uparrow$} & \textbf{AF$\downarrow$} 
& \textbf{AA$\uparrow$} & \textbf{AF$\downarrow$} 
& \textbf{AA$\uparrow$} & \textbf{AF$\downarrow$} 
& \textbf{AA$\uparrow$} & \textbf{AF$\downarrow$} \\
\midrule

Vanilla
& \score{25.57}{0.13} & -- 
& \score{29.28}{0.14} & -- 
& \score{43.78}{0.14} & -- 
& \score{35.85}{0.11} & -- 
& \score{30.84}{0.15} & -- \\

CoT-ICL
& \score{34.37}{0.21} & \score{5.08}{0.13} 
& \score{30.19}{0.24} & \score{6.13}{0.13} 
& \score{46.74}{0.24} & \score{3.29}{0.14} 
& \score{38.74}{0.24} & \score{4.76}{0.11} 
& \score{34.90}{0.20} & \score{5.64}{0.14} \\

SC(CoT)-ICL
& \score{32.54}{0.24} & \score{7.46}{0.11} 
& \score{33.82}{0.22} & \score{6.24}{0.11} 
& \score{50.46}{0.21} & \score{3.40}{0.14} 
& \score{41.51}{0.21} & \score{3.24}{0.14} 
& \score{37.47}{0.22} & \score{5.63}{0.13} \\

Chain
& \score{29.11}{0.14} & -- 
& \score{30.63}{0.16} & -- 
& \score{45.64}{0.17} & -- 
& \score{32.03}{0.16} & -- 
& \score{32.24}{0.17} & -- \\

Tree
& \score{30.40}{0.13} & -- 
& \score{34.29}{0.17} & -- 
& \score{47.12}{0.15} & -- 
& \score{37.81}{0.14} & -- 
& \score{34.17}{0.15} & -- \\

Complete
& \score{32.24}{0.17} & -- 
& \score{35.50}{0.13} & -- 
& \score{41.71}{0.14} & -- 
& \score{39.62}{0.14} & -- 
& \score{36.06}{0.14} & -- \\

\midrule

AFLOW
& \score{28.45}{0.19} & \score{7.52}{0.12} 
& \score{39.46}{0.20} & \score{6.89}{0.14} 
& \score{48.82}{0.21} & \score{6.17}{0.12} 
& \score{40.18}{0.22} & \score{4.72}{0.12} 
& \score{37.76}{0.20} & \score{6.45}{0.10} \\

AFLOW$^\dagger$
& \score{30.21}{0.17} & \score{3.89}{0.12} 
& \score{40.29}{0.20} & \score{4.16}{0.09} 
& \score{54.77}{0.18} & \score{4.22}{0.10} 
& \score{43.68}{0.19} & \score{1.89}{0.12} 
& \score{40.39}{0.20} & \underline{\score{4.12}{0.10}} \\

AgentDropout
& \score{33.15}{0.19} & \score{5.83}{0.13} 
& \score{39.59}{0.23} & \score{5.62}{0.12} 
& \score{48.91}{0.20} & \score{4.03}{0.13} 
& \score{42.94}{0.23} & \score{5.19}{0.14} 
& \score{36.72}{0.21} & \score{8.12}{0.11} \\

AgentDropout$^\dagger$
& \score{36.05}{0.20} & \score{4.07}{0.10} 
& \score{41.17}{0.18} & \score{4.92}{0.10} 
& \score{53.58}{0.19} & \underline{\score{2.28}{0.10}} 
& \score{47.17}{0.18} & \score{1.04}{0.12} 
& \score{40.61}{0.17} & \score{7.48}{0.11} \\

GDesigner
& \score{45.62}{0.18} & \score{6.33}{0.11} 
& \score{42.93}{0.19} & \score{5.14}{0.12} 
& \score{54.46}{0.17} & \score{4.89}{0.11} 
& \score{44.92}{0.18} & \score{5.23}{0.11} 
& \score{41.81}{0.17} & \score{6.95}{0.08} \\

GDesigner$^\dagger$
& \score{47.73}{0.17} & \score{4.20}{0.07} 
& \score{45.52}{0.16} & \underline{\score{3.57}{0.09}} 
& \score{57.13}{0.17} & \score{5.04}{0.11} 
& \score{47.97}{0.16} & \score{1.26}{0.08} 
& \score{44.21}{0.15} & \score{5.15}{0.07} \\

MasRouter
& \score{46.84}{0.20} & \score{4.64}{0.11} 
& \score{43.86}{0.17} & \score{4.02}{0.09} 
& \score{58.03}{0.19} & \score{4.18}{0.11} 
& \score{45.26}{0.17} & \score{3.74}{0.12} 
& \score{45.50}{0.16} & \score{7.48}{0.11} \\

MasRouter$^\dagger$
& \underline{\score{49.16}{0.15}} & \underline{\score{3.01}{0.08}} 
& \underline{\score{46.28}{0.17}} & \score{3.60}{0.07} 
& \textbf{\score{62.09}{0.18}} & \score{3.46}{0.08} 
& \score{48.79}{0.18} & \underline{\score{0.72}{0.10}} 
& \underline{\score{49.55}{0.14}} & \score{5.02}{0.08} \\

ARGDesigner
& \score{47.11}{0.17} & \score{5.36}{0.09} 
& \score{38.94}{0.20} & \score{3.93}{0.11} 
& \score{52.47}{0.18} & \score{4.09}{0.12} 
& \score{46.32}{0.18} & \score{4.27}{0.09} 
& \score{42.18}{0.19} & \score{6.14}{0.12} \\

ARGDesigner$^\dagger$
& \score{48.70}{0.14} & \score{3.16}{0.10} 
& \score{40.40}{0.14} & \score{7.59}{0.11} 
& \score{58.41}{0.15} & \score{2.76}{0.08} 
& \underline{\score{49.32}{0.18}} & \score{1.90}{0.07} 
& \score{49.13}{0.17} & \score{4.83}{0.10} \\

\rowcolor{gray!15}
\textbf{\model}
& \textbf{\score{51.59}{0.12}} & \textbf{\score{0.06}{0.05}} 
& \textbf{\score{47.91}{0.09}} & \textbf{\score{0.57}{0.05}} 
& \underline{\score{61.14}{0.12}} & \textbf{\score{-2.03}{0.06}} 
& \textbf{\score{52.31}{0.09}} & \textbf{\score{-0.55}{0.04}} 
& \textbf{\score{51.67}{0.10}} & \textbf{\score{2.07}{0.06}} \\

\midrule
Rel. Gains
& \textcolor{masfactgreen}{$\uparrow$4.94\%} & \textcolor{masfactgreen}{$\downarrow$98.01\%}
& \textcolor{masfactgreen}{$\uparrow$3.52\%} & \textcolor{masfactgreen}{$\downarrow$84.03\%}
& \textcolor{masfactgreen}{$\downarrow$1.53\%} & \textcolor{masfactgreen}{$\downarrow$189.04\%}
& \textcolor{masfactgreen}{$\uparrow$6.06\%} & \textcolor{masfactgreen}{$\downarrow$176.39\%}
& \textcolor{masfactgreen}{$\uparrow$4.28\%} & \textcolor{masfactgreen}{$\downarrow$49.76\%} \\

\bottomrule
\end{tabular}
}
\end{table*}

%% file: table/Table5_appendix.tex
\begin{table*}[t]
\centering
\scriptsize
\setlength{\tabcolsep}{4.2pt}
\renewcommand{\arraystretch}{1.12}
\caption{
Ablation study on the core designs under MATH 10-shot class-incremental setting. Results include standard deviations over five random seeds.
}
\label{tab:app-ablation}

\begin{tabular}{lcccc}
\toprule
\textbf{Method} & \textbf{AA$\uparrow$} & \textbf{AF$\downarrow$} & \textbf{$\Delta$AA} & \textbf{$\Delta$AF} \\
\midrule

\rowcolor{masfactrow}
\textbf{\model}
& \textbf{59.10{\scriptsize$\pm$0.42}}
& \textbf{-0.41{\scriptsize$\pm$0.18}}
& -- & -- \\

\midrule
w/o Historical Prior
& 54.63{\scriptsize$\pm$0.61}
& 2.36{\scriptsize$\pm$0.27}
& -4.47 & +2.77 \\

\midrule
w/o FGW Alignment
& 55.13{\scriptsize$\pm$0.47}
& 0.76{\scriptsize$\pm$0.22}
& -3.97 & +1.17 \\
Euclidean Alignment
& 54.46{\scriptsize$\pm$0.55}
& 0.94{\scriptsize$\pm$0.25}
& -4.64 & +1.35 \\
GW-only Alignment
& 56.21{\scriptsize$\pm$0.91}
& 0.38{\scriptsize$\pm$0.36}
& -2.89 & +0.79 \\

\midrule
w/o Residual Edit
& 55.01{\scriptsize$\pm$0.39}
& 0.03{\scriptsize$\pm$0.26}
& -4.09 & +0.44 \\
w/o PAC-Bayes Constraint
& 54.71{\scriptsize$\pm$0.29}
& 1.27{\scriptsize$\pm$0.14}
& -4.39 & +1.68 \\
L2 Complexity
& 56.90{\scriptsize$\pm$0.46}
& 0.61{\scriptsize$\pm$0.21}
& -2.20 & +1.02 \\

\bottomrule
\end{tabular}

\end{table*}